\pdfoutput=1

\documentclass[hidelinks]{IEEEtran}

\usepackage{algpseudocode}
\usepackage{amsmath}
\usepackage{amssymb}
\usepackage{amsthm}
\usepackage{booktabs}
\usepackage{cite}
\usepackage{color}
\usepackage{enumitem}
\usepackage{graphicx}
\usepackage{hyperref}
\usepackage{pdfpages}
\usepackage{subfig}

\newtheorem{example}{Example}
\newtheorem{theorem}{Theorem}

\newcommand\fnurl[2]{%
\href{#2}{#1}\footnote{\url{#2}}%
}

\IEEEoverridecommandlockouts
\IEEEpubid{\begin{minipage}[t]{1.1\textwidth}\ \\[10pt]
        \centering{\fontsize{6.5}{7} \selectfont \copyright 2018 IEEE. Personal use of this material is permitted. Permission from IEEE must be obtained for all other uses, in any current or future media, including reprinting/republishing this material for advertising or promotional purposes, creating new collective works, for resale or redistribution to servers or lists, or reuse of any copyrighted component of this work in other works. Citation information: DOI 10.1109/TFUZZ.2019.2900856, IEEE Transactions on Fuzzy Systems}
\end{minipage}}

\begin{document}

\title{CFM-BD: a distributed rule induction algorithm for building Compact Fuzzy Models in Big Data classification problems}

\author{Mikel Elkano, 
		Jose Sanz,
        Edurne Barrenechea,
        Humberto~Bustince,
        Mikel Galar
\thanks{This work has been supported by the Spanish Ministry of Economy and Competitiveness under the project TIN2016-77356-P (MINECO, AEI/FEDER, UE).}
\thanks{Mikel Elkano, Jose Sanz, Edurne Barrenechea, Humberto Bustince, and Mikel Galar are with the Institute of Smart Cities and the Department of Statistics, Computer Science and Mathematics, Universidad P\'ublica de Navarra, 31006 Pamplona, Spain. Humberto Bustince is also with King Abdulaziz University, Jedda, Saudi Arabia. E-mails: \{mikel.elkano, joseantonio.sanz, edurne.barrenechea, bustince, mikel.galar\}@unavarra.es}
}

\maketitle

\begin{abstract}
Interpretability has always been a major concern for fuzzy rule-based classifiers. The usage of human-readable models allows them to explain the reasoning behind their predictions and decisions. However, when it comes to Big Data classification problems, fuzzy rule-based classifiers have not been able to maintain the good trade-off between accuracy and interpretability that has characterized these techniques in non-Big Data environments. The most accurate methods build too complex models composed of a large number of rules and fuzzy sets, while those approaches focusing on interpretability do not provide state-of-the-art discrimination capabilities. In this paper, we propose a new distributed learning algorithm named CFM-BD to construct accurate and compact fuzzy rule-based classification systems for Big Data. This method has been specifically designed from scratch for Big Data problems and does not adapt or extend any existing algorithm. The proposed learning process consists of three stages: 1) pre-processing based on the probability integral transform theorem; 2) rule induction inspired by CHI-BD and Apriori algorithms; 3) rule selection by means of a global evolutionary optimization. We conducted a complete empirical study to test the performance of our approach in terms of accuracy, complexity, and runtime. The results obtained were compared and contrasted with four state-of-the-art fuzzy classifiers for Big Data (FBDT, FMDT, Chi-Spark-RS, and CHI-BD). According to this study, CFM-BD is able to provide competitive discrimination capabilities using significantly simpler models composed of a few rules of less than 3 antecedents, employing 5 linguistic labels for all variables.
\end{abstract}


\IEEEpeerreviewmaketitle

\section{Introduction}\label{sec:intro}

\IEEEPARstart{F}UZZY Rule-Based Classification Systems (FRBCSs) are powerful machine learning algorithms that provide interpretable and accurate models described by human-readable linguistic labels~\cite{Ishibuchi2004}. The main feature that makes FRBCSs stand out from other types of solutions is the ability to explain how outputs are inferred from inputs. Due to this valuable reasoning, these systems have been widely used in applications such as bioinformatics~\cite{Sarkar2016}, medical problems~\cite{Sanz2013b}, software fault prediction~\cite{Singh2017}, anomaly intrusion detection~\cite{Tsang2007}, financial applications~\cite{Antonelli2017}, image processing~\cite{Nakashima2007}, and traffic congestion prediction~\cite{Zhang2014}, among others.

However, the construction of interpretable models usually involves computationally intensive learning algorithms that require long runtimes. As a consequence, state-of-the-art FRBCSs have serious difficulties dealing with large-scale datasets. Given the increasing amount of information in the Digital Age, which is doubling every two years according to the study carried out by the International Data Corporation (IDC)~\cite{Gantz2012}, the design of new scalable solutions presents many challenges. Some of the latest publicly available Big Data classification problems, such as \fnurl{HIGGS or HEPMASS}{http://archive.ics.uci.edu/ml/datasets.html}, do not fit into the main memory of standard computers. Therefore, state-of-the-art sequential learning algorithms are not able to handle the whole training set on a single computer. Moreover, even if such quantity of data could be stored in an 8-GB RAM memory, the training process would probably lead to unacceptable runtimes.

In order to overcome these challenges, many researchers started to adapt well-known machine learning techniques to \emph{distributed computing} paradigms such as MapReduce~\cite{Dean2008,Owen2011,Meng2016}. This methodology rapidly became very popular as a result of the development of open-source frameworks such as \fnurl{Apache Hadoop}{http://hadoop.apache.org} and \fnurl{Apache Spark}{https://spark.apache.org}. In the last few years, a number of distributed FRBCs based on either Hadoop or Spark have been proposed~\cite{Ducange2015,Elkano2017,Fernandez2016b,Fernandez2017,Ferranti2017,Garcia-Vico2017,Lopez2015,Pulgar-Rubio2017,Segatori2018a,Segatori2018b}. Although great progress has been made, most contributions do not achieve state-of-the-art results in terms of both accuracy and interpretability. Some of them perform several local optimization or learning processes to obtain an approximate global solution~\cite{Fernandez2016b,Fernandez2017,Lopez2015,Pulgar-Rubio2017}, missing important patterns that could only be extracted when the training set is treated as a whole. Other methods produce too complex models that affect interpretability, mainly due to a large number of rules or fuzzy sets~\cite{Elkano2017,Segatori2018a,Segatori2018b}. There are also other contributions that, from our point of view, do not include enough Big Data problems to assess their quality in the corresponding experimental study~\cite{Ducange2015,Garcia-Vico2017}.

In this work, we propose a new distributed FRBCS named CFM-BD to build compact and accurate models for Big Data classification problems. Our objective is to generate rule bases containing a few (short) fuzzy rules using a small fixed number of fuzzy sets per variable, while achieving state-of-the-art classification performance. The proposed algorithm consists of three sequential stages:
\begin{enumerate}
\item \emph{Pre-processing and partitioning}. Training data is transformed into a uniform distribution by applying the \emph{probability integral transform theorem}~\cite{Angus1994,Quesenberry2004}. Next, the fuzzy sets are uniformly distributed in the new transformed space. In this manner, the partitions fit the actual distribution of the training data and can be safely recovered in the original space by making use of the \emph{inverse cumulative distribution function} or \emph{quantile function}~\cite{Nair2013}.
\item \emph{Rule induction process}. Rules are constructed by a novel learning algorithm inspired by CHI-BD~\cite{Elkano2017} and Apriori~\cite{Agrawal1994} algorithms. First, the most frequent itemsets are extracted from the initial rules generated by CHI-BD and a pruning process is then carried out. Next, the itemsets are converted into candidate rules and a filtering and pruning process is performed to select the rules with the greatest discrimination capability.
\item \emph{Evolutionary rule selection}. We implement our own distributed version of the CHC evolutionary algorithm~\cite{Eshelman1991} to optimize the rule base by means of rule selection. To the best of our knowledge, this is the first distributed solution for global evolutionary rule selection.
\end{enumerate}
We must remark that all the stages process the whole dataset in a distributed fashion and the model obtained is not affected by the distribution of the partitions and the degree of parallelism. The full source code was written in \fnurl{Scala}{https://www.scala-lang.org} 2.11 on top of Apache Spark 2.0.2 and is publicly available at \fnurl{GitHub}{https://github.com/melkano/cfm-bd} under the GPL license.

In order to assess the performance of our method, we carried out an empirical study using 6 Big Data classification problems available at UCI~\cite{Lichman2013} and \fnurl{OpenML}{https://www.openml.org/search?type=data} repositories. Accuracy, complexity, and runtimes were analyzed and compared with the results obtained by three state-of-the-art fuzzy classifiers, i.e., CHI-BD~\cite{Elkano2017}, Chi-Spark-RS~\cite{Fernandez2017}, and FMDT/FBDT\cite{Segatori2018b}. Additionally, the scalability of our approach was measured with three well-known metrics used to evaluate distributed systems, i.e., \emph{speedup}, \emph{sizeup}, and \emph{scaleup}~\cite{DeWitt1992,Jogalekar2000}. The experimental results show that CFM-BD is able to deal with large-scale datasets and achieve competitive accuracy rates while providing simpler models than the aforementioned algorithms.

This paper is organized as follows. Section \ref{sec:preliminaries} includes the basics of FRBCSs and Apache Hadoop/Spark and presents some related work. In Section \ref{sec:cfm-bd}, we introduce the proposed distributed FRBCS for Big Data classification problems. The experimental framework is described in Section \ref{sec:experimental-framework} and the analysis of the empirical results is presented in Section \ref{sec:experimental-study}. Finally, Section \ref{sec:conclusions} concludes this paper.

\section{Preliminaries}\label{sec:preliminaries}

In this section we briefly describe some basic concepts and frameworks that are directly related to our proposal. First, we explain the basics of Fuzzy Rule-Based Classification Systems (Section \ref{ssec:frbcs}). Next, we introduce two popular frameworks used to handle Big Data environments called Apache Hadoop and Apache Spark (Section \ref{ssec:hadoop-spark}). Finally, we present and discuss some recent related work (Section \ref{ssec:related}).

\subsection{Fuzzy Rule-Based Classification Systems} \label{ssec:frbcs}

Fuzzy Rule-Based Classification Systems (FRBCSs) are well-known models that achieve a good trade-off between classification accuracy and interpretability. These systems provide an interpretable rule base containing human-readable rules composed of linguistic labels~\cite{Ishibuchi2004}.

The two main components of FRBCSs are described hereafter.

\begin{enumerate}
\setlength{\itemsep}{-\parsep}
\item \emph{Knowledge base (KB)}: It is composed of both the rule base (RB) and the database (DB), where the rules and membership functions used to model the linguistic labels are stored, respectively.
\item \emph{Fuzzy Reasoning Method (FRM)}: This is the mechanism used to classify examples employing the information stored in the KB.
\end{enumerate}

In order to generate the KB, a fuzzy rule learning algorithm is applied using a training set $TR$ composed of $N$ labeled examples $x_i = (x_{i1}, \ldots, x_{iF})$ with $i \in \left\lbrace1,\ldots,N\right\rbrace$, where $x_{if}$ is the value of the $f$-th feature $(f \in \left\lbrace1,2,\ldots,F\right\rbrace)$ of the $i$-th training example. Each example belongs to a class $y_i \in \mathbb{C} = \{C_1,C_2,...,C_M\}$, where $M$ is the number of classes in the problem. 

The RB is composed of a set of rules having the following structure:
\begin{equation}
\label{eq:fuzzy-rule}
\begin{split}
 \mbox{Rule } R_j: &\mbox{ If } x_1 \mbox{ is } A_{j1} \mbox{ and } \ldots \mbox{ and } x_F \mbox{ is } A_{jF}  \\ &\mbox{ then Class = } C_j \mbox{ with } RW_j
 \end{split}
\end{equation}
where $R_j$ is the label of the $j$-th rule, $x = (x_{1}, \ldots, x_{F})$ is an $F$-dimensional pattern vector that represents the example, $A_{jf}$ is a linguistic label modeled by a membership function, $C_j$ is the class label and $RW_{j}$ is the rule weight. In some cases rules might contain \emph{don't care} linguistic labels making the classifier to ignore the corresponding attribute value. These labels can simply be removed from the RB, leading to variable rule lengths. Regarding the rule weight computation, in this work we use an adaptation of the well-known Penalized Certainty Factor (PFC) method~\cite{Ishibuchi2005} called Penalized Cost-Sensitive Certainty Factor (PCF-CS). This method minimizes the impact of the frequency of each class on the learning process by applying the following formula:
\begin{equation}\label{eq:rule-weight}
RW_j = \frac{matchClass - matchNotClass}{matchClass + matchNotClass},
\end{equation}
\noindent where 
\begin{equation*}
\begin{split}
matchClass = \sum_{x_i \in Class C_j}\mu_{A_j}(x_i)\cdot cost(y_i) \\
matchNotClass = \sum_{x_i \not\in Class C_j}\mu_{A_j}(x_i)\cdot cost(y_i).
\end{split}
\end{equation*}
\noindent $cost(y_i)$ is the misclassification cost associated with the class $y_i$ and $\mu_{A_j}(x_i)$ is the matching degree between the example $x_i$ and the antecedent part of the rule $R_j$. The original PFC formula is recovered when the same cost ($cost(y_i)=1$) is given to all classes. Although class costs were originally considered for binary classification problems, we have adapted their computation to multi-class problems as follows:
\begin{equation}\label{eq:class-cost}
cost(y_i) = \frac{\displaystyle \underset{m=1,\ldots,M}{max} \left(count(y_m)\right)}{count(y_i)},
\end{equation}
\noindent where $count(y_i)$ is the number of examples belonging to the class $y_i$. As for the matching degree $\mu_{A_j}(x_i)$, it is defined as:
\begin{equation}\label{eq:matching-degree}
\mu_{A_j}(x_i) = \prod\limits_{f=1}^F \mu_{A_{jf}}(x_{if}),
\end{equation}
\noindent where $\mu_{A_{jf}}(x_{if})$ is the membership degree of the value $x_{if}$ to the fuzzy set $A_{jf}$ of the rule $R_j$. If $A_{jf}$ is marked as \emph{don't care}, the membership degree is set to 1.

In order to classify a new example $x_i$, the classifier runs an FRM composed of the following steps.

\begin{enumerate}
\item \emph{Matching degree}. The strength of activation of the antecedent part of all rules in the rule base for the example $x_i$ is computed (Eq. (\ref{eq:matching-degree})).
\item \emph{Association  degree}. The association degree of the example $x_i$ with each rule in the rule base is computed.
\begin{equation}\label{eq:association-degree}
b_j(x_i) = \mu_{A_j}(x_i) \cdot RW_j
\end{equation}
\item \emph{Classification}. The final prediction is made based on the association degrees. In this work we use the \emph{winning rule} method~\cite{Cordon1999}, which predicts the class of the rule with the highest association degree:
\begin{equation}\label{eq:winning-rule}
class = \displaystyle arg\ \underset{m=1,\ldots,M}{max} \left( \underset{R_j \in RB; \; C_j=m}{max}b_j(x_i) \right)
\end{equation}
%The two most common methods~\cite{Cordon1999} are:
%\begin{itemize}
%\item \emph{Winning rule}. The class of the rule with the highest association degree is predicted.
%\begin{equation}\label{eq:winning-rule}
%class = \displaystyle arg\ \underset{m=1,\ldots,M}{max} \left( \underset{R_j \in RB; \; C_j=m}{max}b_j(x_i) \right)
%\end{equation}
%\item \emph{Additive combination}. The association degrees of all classes are summed up, obtaining the \emph{confidence} of each class. Finally, the class having the highest confidence is predicted.
%\begin{equation}\label{eq:additive-combination}
%\begin{split}
%&conf_m(x_i) = \displaystyle \sum_{R_j \in RB; \; C_j=m}{b_j(x_i)},\mbox{ }m=1,\ldots,M
%\\
%&class(x_i) = \displaystyle arg\ \underset{m = 1,\ldots,M}{max}\left(conf_m(x_i)\right)
%\end{split}
%\end{equation}
%\end{itemize}
\end{enumerate}

\subsection{Apache Hadoop and Apache Spark} \label{ssec:hadoop-spark}
%The fundamental idea behind this strategy is to distribute the training data across multiple computing nodes that concurrently process a subset (partition) of the whole dataset. Next, the partial results produced on these nodes are aggregated to obtain the final result.
In the last few years, distributed computing has become very popular in the machine learning community thanks to open-source frameworks such as \fnurl{Apache Hadoop}{http://hadoop.apache.org} and \fnurl{Apache Spark}{https://spark.apache.org}. These frameworks provide a transparent distributed system that allows the user to focus only on data processing. The core of Hadoop consists of a distributed file system based on the Google File System~\cite{Ghemawat2003} called \emph{Hadoop Distributed File System} (HDFS) (storage layer) and an implementation of the MapReduce paradigm~\cite{Dean2008} (processing layer).

\begin{figure*}[!htbp]
\centering
\resizebox{\textwidth}{!}{\fbox{\includegraphics{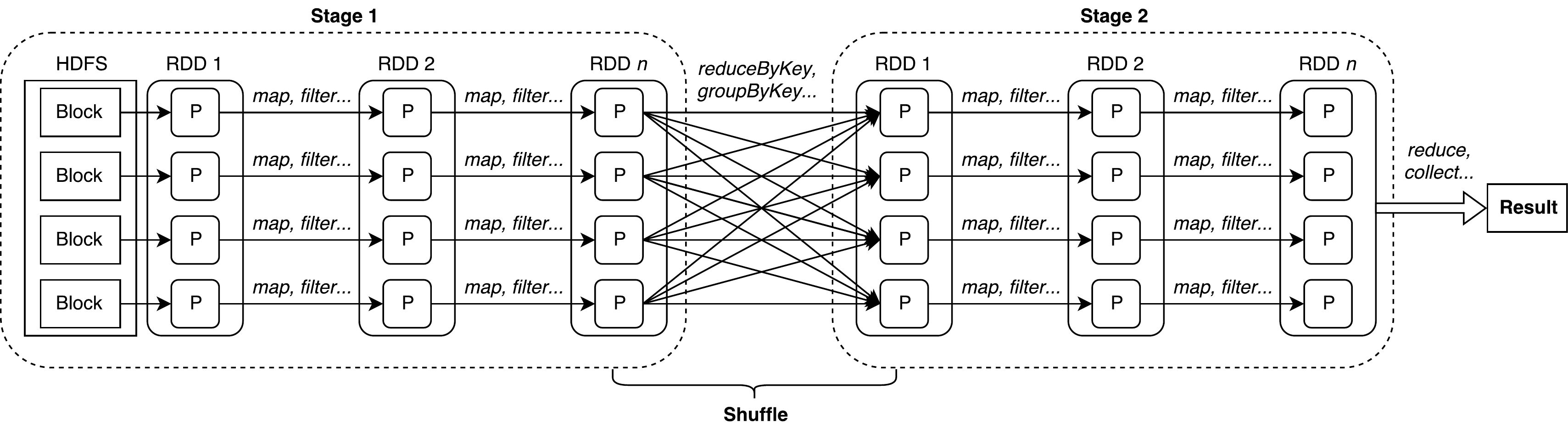}}}\caption{Spark's data flow (P = Partition).}\label{fig:spark-diagram}
\end{figure*}

Spark was introduced as a generalization and an extension of the MapReduce paradigm. It has seemingly unseated Hadoop thanks to the so-called Directed Acyclic Graphs (DAGs) and the in-memory computing feature, which minimize the latency of multi-stage data flows that require multiple MapReduce jobs. Spark is built around the concept of \emph{Resilient Distributed Datasets} (RDDs)~\cite{Zaharia2012}, which represent distributed immutable data (partitioned data) and lazily evaluated operations (\emph{transformations}). The execution of a user-defined algorithm consists of a sequence of \emph{stages} composed of a number of transformations that are split into \emph{tasks}. One stage consists only of transformations that do not require any shuffling/repartitioning process (e.g., \emph{map} and \emph{filter} operations). Tasks are executed by the so-called \emph{executors}, which represent independent processes in the Java Virtual Machine (JVM) of a \emph{worker} node. Finally, the result of all transformations is obtained by calling an \emph{action} that computes and returns the result to the \emph{driver} node. This data flow (Fig. \ref{fig:spark-diagram}) allows the user to run an indefinite number of MapReduce jobs within the same main program, supporting a much wider variety of algorithms and methods than Hadoop.

\subsection{Related work}\label{ssec:related}

To the best of our knowledge, barely a dozen interpretable fuzzy methods have been proposed to deal with Big Data classification problems~\cite{Ducange2015,Elkano2017,Fernandez2016b,Fernandez2017,Ferranti2017,Gamez2016,Garcia-Vico2017,Lopez2015,Pulgar-Rubio2017,Romero-Zaliz2017,Segatori2018a,Segatori2018b}. Although there exist more contributions that use fuzzy logic for classification \cite{Deng2015,Li2017}, their scalability has not been proven in Big Data problems, and thus they are out of the scope of this paper. Given the success of fuzzy classifiers in a wide range of fields~\cite{Sarkar2016,Sanz2013b,Singh2017,Tsang2007,Antonelli2017,Nakashima2007,Zhang2014}, designing scalable solutions seems worth the effort. In~\cite{Fernandez2016a}, the authors provide an overview of the progress and opportunities of fuzzy logic in Big Data environments.

In general, the aforementioned solutions are based on either incremental learning~\cite{Gamez2016,Romero-Zaliz2017} or distributed computing~\cite{Ducange2015,Elkano2017,Fernandez2016b,Fernandez2017,Ferranti2017,Garcia-Vico2017,Lopez2015,Pulgar-Rubio2017,Segatori2018a,Segatori2018b}. In the former case, training data is divided into several subsets called \emph{episodes} that sequentially feed a classifier that incrementally learns from input data, including the knowledge acquired in previous episodes. The advantage of this approach is that it does not require a computing cluster to run the algorithm, since each episode is equivalent to a small-data classification problem in terms of computational cost. Regarding distributed solutions, the learning process is carried out by distributing training data across several computing nodes that perform partial computations in order to get the final model. The main difference between this strategy and incremental learning is that the former runs a single learning process in parallel using the whole training set, while the latter carries out several learning stages in a sequential fashion. Therefore, it is clear that the drawback of distributed approaches is the need for several computing nodes. However, incremental learning might miss important knowledge coming from inter-relations among data from different episodes, since it lacks the overview of the whole training set. In this work, we focus on the design of distributed methods.

Distributed learning algorithms might, in turn, tackle classification problems either by decomposing the original training data into several local sub-problems~\cite{Lopez2015,Fernandez2016b,Fernandez2017} or by performing a global distributed learning process~\cite{Elkano2017,Ferranti2017,Garcia-Vico2017,Segatori2018a,Segatori2018b}, or even by combining these two approaches~\cite{Ducange2015,Pulgar-Rubio2017}. In the former case, an independent local model is concurrently built in each subset (\emph{chunk}) of data, so that the final classifier is obtained by aggregating all these models. In this manner, one could apply a well-known non-distributed algorithm to train each local model. However, similarly to incremental learning, the learning process becomes strongly dependent on the distribution of subsets and might miss important information available only when training data is treated as a whole. Regarding global distributed learning algorithms, the difficulty of parallelizing the training phase across several computing units is the main drawback.

Different strategies have been applied to obtain human-readable fuzzy models in Big Data classification problems, including fuzzy versions of decision trees (FDTs)~\cite{Ferranti2017,Segatori2018b}, sub-group discovery (SD)~\cite{Pulgar-Rubio2017}, associative classifiers (FACs)~\cite{Ducange2015,Segatori2018a}, emerging patterns mining (EPM)~\cite{Garcia-Vico2017}, and rule-based classifiers (FRBCs)~\cite{Elkano2017,Fernandez2016b,Fernandez2017,Ferranti2017,Lopez2015}. In~\cite{Ferranti2017}, a distributed version of C4.5 is used to extract a candidate rule base that is optimized by an evolutionary algorithm. Segatori et al. proposed a distributed FDT that exploits the classical Decision Tree implementation in \fnurl{Spark MLlib}{http://spark.apache.org/mllib}, extending the learning scheme by employing fuzzy information gain based on fuzzy entropy~\cite{Segatori2018b}. A new algorithm for SD called MEFASD-BD was presented in~\cite{Pulgar-Rubio2017}, which makes use of an evolutionary fuzzy system to extract fuzzy rules describing subgroups for each partition, though the quality of each solution is measured on the whole training set. Fuzzy logic was also used for EPM in Big Data by Garc\'ia-Vico et al.~\cite{Garcia-Vico2017}, applying a global evolutionary fuzzy system that employs the entire training set. Finally, different distributed versions of both FACs and FRBCs were proposed in~\cite{Ducange2015,Elkano2017,Lopez2015,Fernandez2016b,Fernandez2017,Segatori2018a}.

However, the above-mentioned algorithms sacrifice either interpretability for classification accuracy or viceversa. Some algorithms focus on the accuracy and tend to generate too complex models having large amounts of rules~\cite{Elkano2017,Fernandez2016b,Lopez2015}, excessive rule lengths~\cite{Elkano2017,Fernandez2016b,Fernandez2017,Lopez2015}, or a high number of fuzzy sets (linguistic labels)~\cite{Segatori2018a,Segatori2018b}. On the other hand, those algorithms that optimize the interpretability of the model are not able to achieve state-of-the-art results in terms of accuracy~\cite{Ferranti2017}. There are also other contributions that, from our point of view, do not consider enough datasets to assess these aspects in Big Data environments~\cite{Ducange2015,Garcia-Vico2017,Pulgar-Rubio2017}.

\section{CFM-BD: a new distributed fuzzy rule induction algorithm for big data}\label{sec:cfm-bd}

In this work we present CFM-BD, a new distributed fuzzy rule induction algorithm specifically designed for Big Data classification problems. The motivation behind our proposal is to build compact fuzzy models achieving state-of-the-art results in terms of both classification performance and interpretability.

In order to overcome this challenge, we propose a learning algorithm composed of three stages:
\begin{enumerate}
\item Pre-processing and partitioning (Section \ref{ssec:partitioning}).
\item Rule induction process (Section \ref{ssec:rule-induction}).
\item Evolutionary rule selection (Section \ref{ssec:evolutionary}).
\end{enumerate}
%\begin{enumerate}
%\item \emph{Pre-processing and partitioning}. An adaptive fuzzy partitioning process is carried out by applying a novel pre-processing method based on the \emph{probability integral transform theorem}~\cite{Angus1994,Quesenberry2004}.
%\item \emph{Rule induction process}. Rules are constructed by a novel learning algorithm inspired by CHI-BD~\cite{Elkano2017} and Apriori~\cite{Agrawal1994} algorithms.
%\item \emph{Evolutionary rule selection}. A rule selection process is conducted by a new distributed version of the CHC evolutionary algorithm~\cite{Eshelman1991}.
%\end{enumerate}
We must remark that all stages are conducted by distributed processes that employ the whole training set. Therefore, no approximations are introduced throughout the whole execution of the algorithm. This feature allows the user to obtain exactly the same model regardless of the distribution of data partitions and the parallelization degree used for the execution.

The full source code was written in \fnurl{Scala}{https://www.scala-lang.org} 2.11 on top of Apache Spark 2.0.2 and is publicly available at \fnurl{GitHub}{https://github.com/melkano/cfm-bd} under the GPL license.

\subsection{Pre-processing and partitioning}\label{ssec:partitioning}

The goal of this stage is to build fuzzy sets (linguistic labels) that fit the real distribution of the training data while keeping the number of fuzzy sets per variable constant (e. g., \emph{low}, \emph{medium}, \emph{high}). This process is divided into two parts:
\begin{itemize}
\item Pre-processing: the original distribution of the training data is transformed into a uniform distribution. This transformation applies the \emph{probability integral transform theorem}~\cite{Angus1994,Quesenberry2004}, described in Theorem \ref{thm:integral-transform}. This theorem implies that any dataset can be transformed into a new dataset where all the variables follow a uniform distribution, regardless of the original distribution.
\begin{theorem}\label{thm:integral-transform}
If $X$ is a continuous random variable with cumulative distribution function (CDF) $F_X(x)$ and if $Y=F_X(X)$, then $Y$ is a uniform random variable on the interval [0,1].
\end{theorem}
\begin{proof}
Suppose that $Y=g(X)$ is a function of $X$ where $g$ is differentiable and strictly increasing. Thus, its inverse $g^{-1}$ uniquely exists. The CDF of $Y$ can be derived using
\begin{equation*}
\begin{split}
F_Y(y)&=Prob\left(Y \leq y\right)=Prob\left(X \leq g^{-1}(y)\right) \\
&=F_X\left(g^{-1}(y)\right)
\end{split}
\end{equation*}
and its density is given by
\begin{equation*}
\begin{split}
f_Y(y)&=\frac{d}{dy}F_Y(y)=\frac{d}{dy}F_X(g^{-1}(y)) \\
&=f_X(g^{-1}(y)) \cdot \frac{d}{dy}g^{-1}(y).
\end{split}
\end{equation*}
\noindent This procedure is called the CDF technique and allows the distribution of $Y$ to be derived as follows:
\begin{equation*}
\begin{split}
F_Y(y)&=Prob\left(Y \leq y\right)=Prob\left(X \leq F_X^{-1}(y)\right) \\
&=F_X\left(F_X^{-1}(y)\right)=y
\end{split}
\end{equation*}
\end{proof}
However, since the original distribution of the training set is unknown, we cannot compute the exact CDF. Instead, we propose computing the $q$-quantiles of the training set and compute an approximate CDF. To this end, for each variable, all the values are sorted and each quantile is extracted. If $q$ is smaller than the number of examples in the training set, the CDF of a certain value is linearly interpolated on the interval [$Q_{i-1}$, $Q_i$], $Q_i$ being the first quantile greater than the value. If the value is smaller than the first quantile ($Q_1$) or greater than the last quantile ($Q_{q-1}$), the CDF is 0 or 1, respectively. Of course, the transformation of the testing set is performed by interpolating the CDF using the quantiles extracted from the training set.
\item Partitioning: the fuzzy sets are built using triangular membership functions and uniformly distributed across the interval [0,1] in the new transformed space. It is worth noting that the definition of every single fuzzy set in the original space can be recovered by applying the \emph{inverse cumulative distribution function} or \emph{quantile function}~\cite{Nair2013}. In this case, for every point defining the triangular membership function, we would linearly interpolate the corresponding value between the two closest quantiles by computing the inverse of the linear function used to compute the CDF.
\end{itemize}

\begin{figure}[!htbp]
\centering
\subfloat[Transformed space of every variable]{\includegraphics[width=\linewidth]{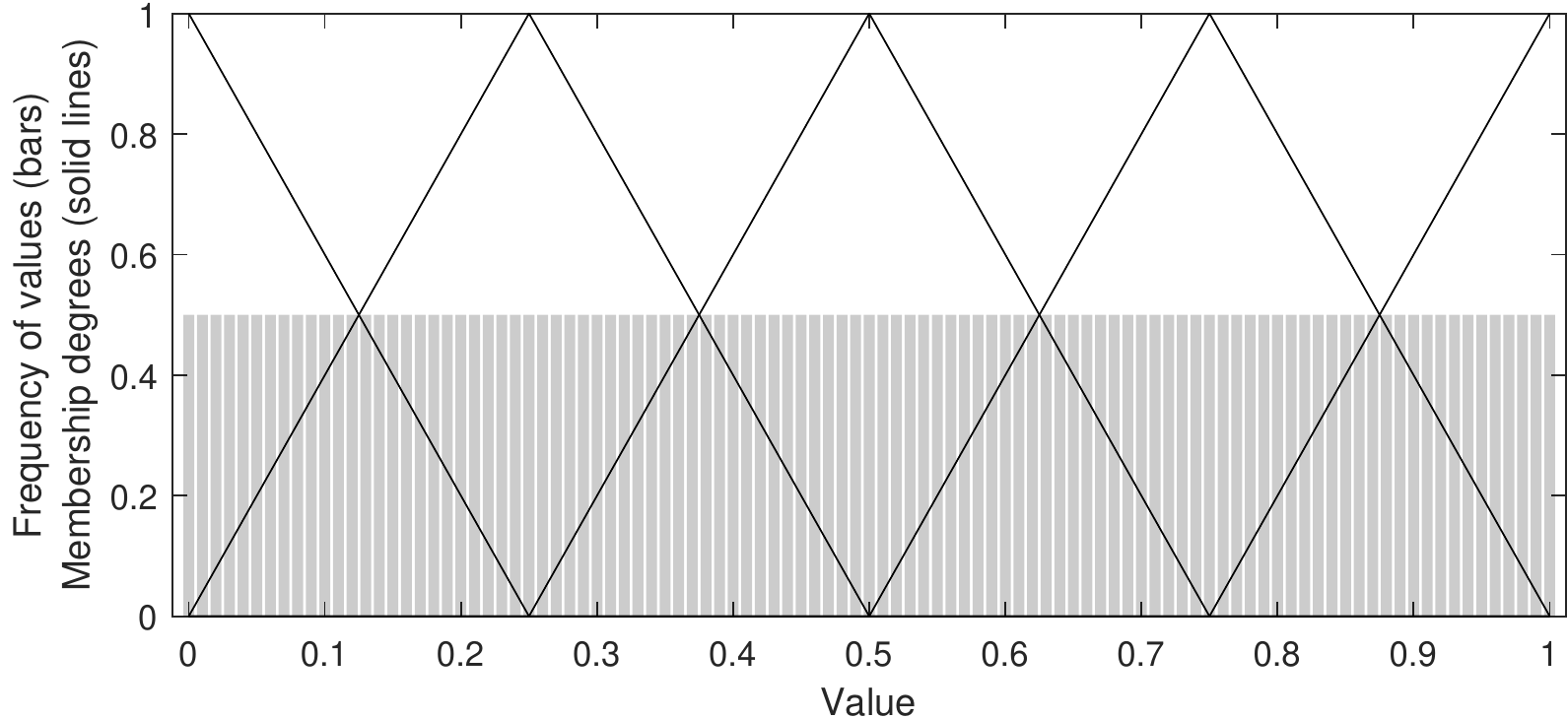}}\\\vspace{-10pt}
\subfloat[Original space of the variable \emph{lepton\_1\_eta}]{\includegraphics[width=\linewidth]{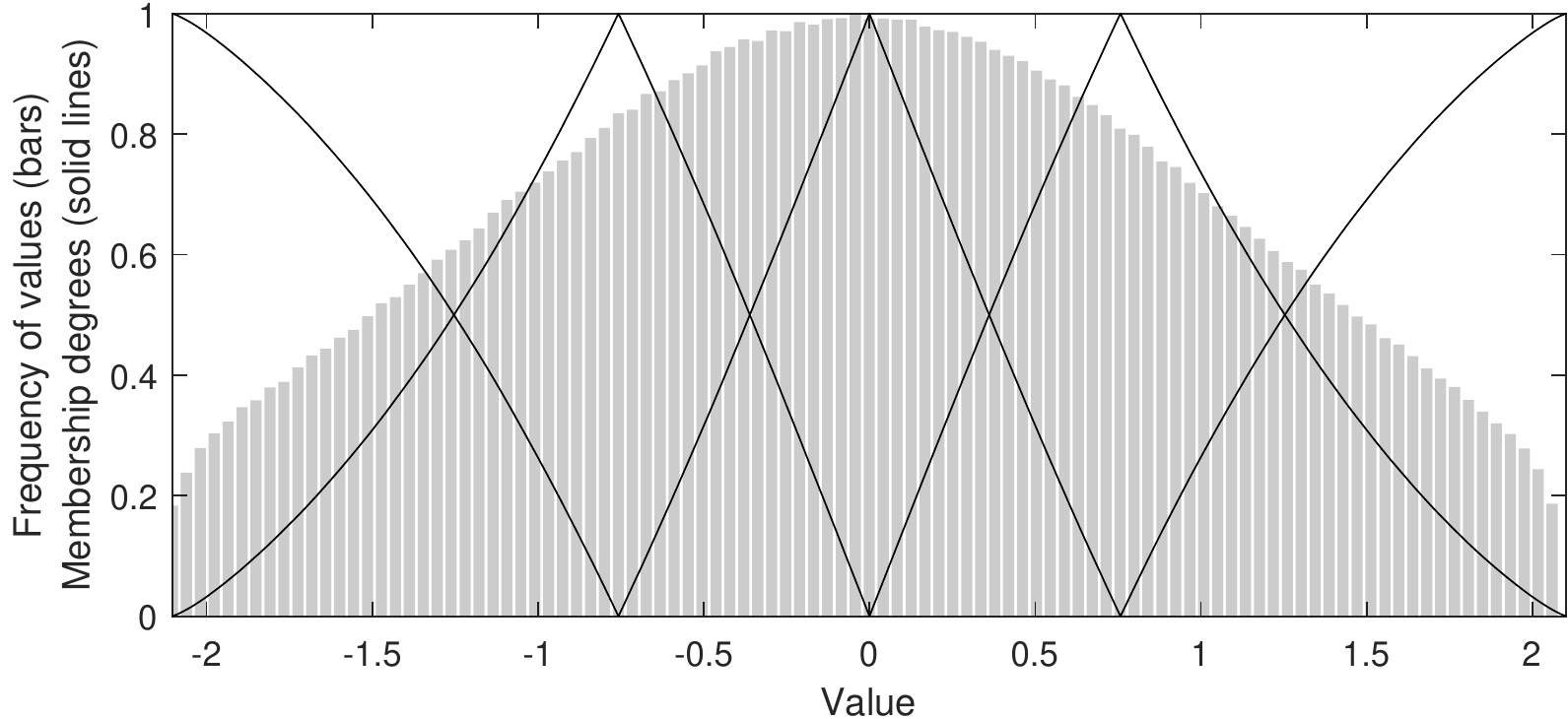}}\\\vspace{-10pt}
\subfloat[Original space of the variable \emph{R}]{\includegraphics[width=\linewidth]{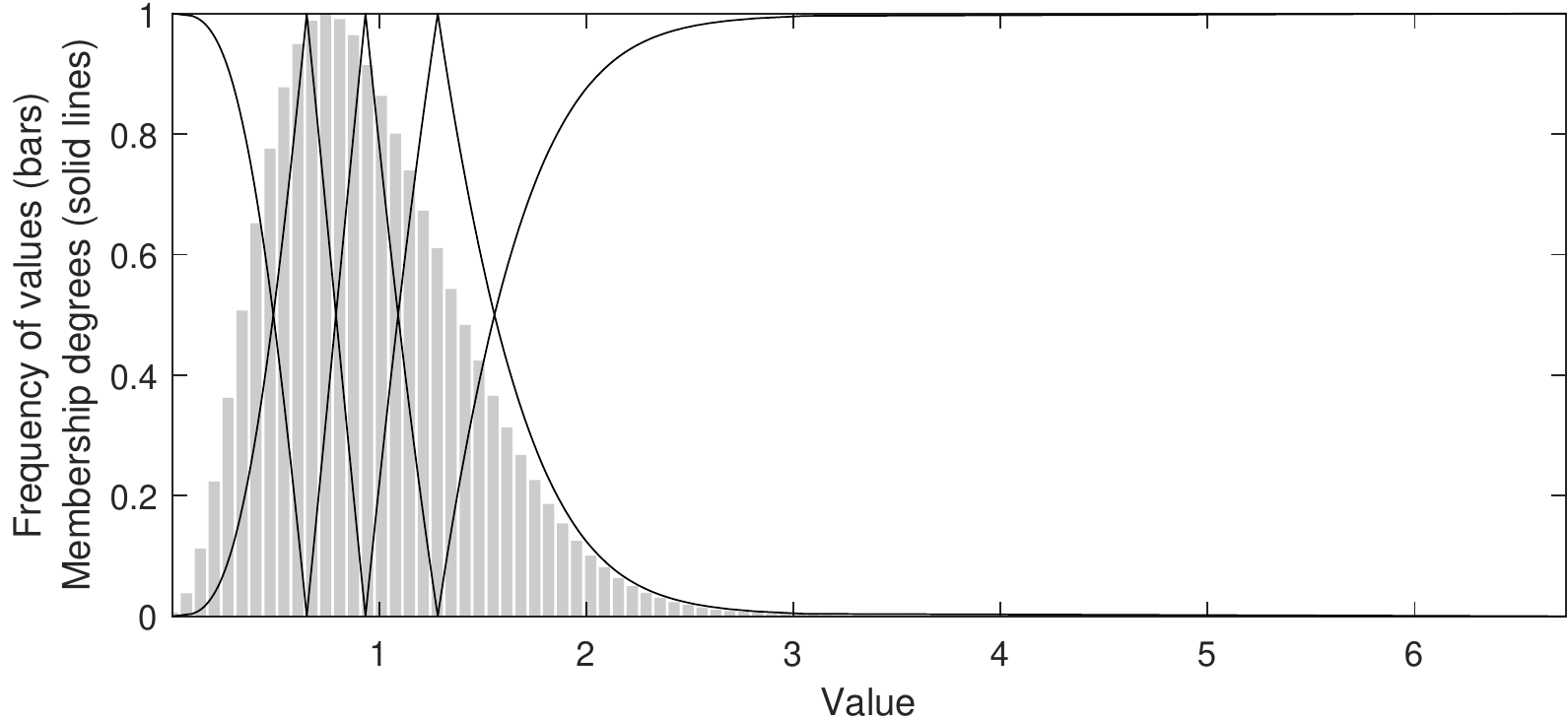}}
\caption{Fuzzy sets built for \emph{lepton\_1\_eta} and \emph{R} on SUSY.}
\label{fig:partitioning}
\end{figure}

Fig. \ref{fig:partitioning} shows an illustrative example of how fuzzy sets are distributed in the original and transformed spaces of the variables \emph{lepton\_1\_eta} and \emph{R} of \fnurl{SUSY}{https://archive.ics.uci.edu/ml/datasets/SUSY}. Solid lines and bar plots represent the membership functions of the fuzzy sets and the original distribution of the variables, respectively.

This stage plays an important role in our method's robustness and is key to building accurate fuzzy rules. We provide additional experiments in this respect in the supplementary material accompanying this paper.

\subsection{Rule induction process}\label{ssec:rule-induction}

After the training data has been pre-processed and the fuzzy sets have been built, the rule base is constructed by applying a new rule induction algorithm specifically designed for Big Data. This process consists of two sequential stages that are inspired by some of the concepts introduced in CHI-BD~\cite{Elkano2017} and Apriori~\cite{Agrawal1994} algorithms.

\subsubsection{Search for the most promising itemsets}\label{sssec:promising-itemsets}

In this work, we consider all fuzzy sets (linguistic labels) and nominal values as \emph{items}. When some of these items (one or more) appear together in a given transaction (example), they form an \emph{itemset}. In order to find out the most promising itemsets in terms of discrimination capability, a procedure composed of three steps is applied.
\begin{enumerate}[label=(\alph*),leftmargin=2.5\parindent]
\item Discretization of the examples: all the itemsets that are present in the training set are extracted by making use of the rule generation process performed by the CHI-BD algorithm~\cite{Elkano2017}. This procedure consists in discretizing all the examples by computing the membership degree of each value to all the fuzzy sets of the corresponding variable. That is, each value is replaced with the fuzzy set leading to the highest membership degree. In case of nominal values, no discretization is conducted. Example \ref{ex:discretization} illustrates a discretized example, where the subscript indicates the variable that the fuzzy set corresponds to.
\begin{example}\label{ex:discretization}
\normalfont
Given the following example in the transformed input space:\vspace{3pt}\\
$0.15, 0.82, 0.51,$\vspace{3pt}\\
and considering three uniformly distributed triangular membership functions for each variable, it could be discretized as:\vspace{3pt}\\
$Low_1, High_2, Medium_3$
\end{example}
In this manner, an $F$-dimensional example is transformed into an itemset of $F$ items. After discretizing the example, all the possible subsets of items contained in the itemset are generated. The maximum cardinality of these subsets is set by the user and determines the maximum length of the rules that will be built in subsequent stages. In this work we have set this value to 3 in order to achieve a good trade-off between model complexity and discrimination capability, based on the results obtained in \cite{AlcalaFdez2011}. Example \ref{ex:itemsets} shows the itemsets extracted from Example \ref{ex:discretization}.
\begin{example}\label{ex:itemsets}
\normalfont
Given the following discretized example:\vspace{3pt}\\
$Low_1, High_2, Medium_3,$\vspace{3pt}\\
all the possible itemsets are:\vspace{3pt}\\
$\{Low_1\}, \{High_2\}, \{Medium_3\},$\\
$\{Low_1, High_2\}, \{Low_1, Medium_3\},$\\
$\{High_2, Medium_3\},$\\
$\{Low_1, High_2, Medium_3\}$
\end{example}
\item Search for frequent itemsets: the \emph{support} of each itemset is computed and only those having a minimum support are kept. In the original Apriori algorithm~\cite{Agrawal1994}, the support of an itemset is the number of times that the itemset appears in the training set. In this work, the support of an itemset $I$ is redefined as
\begin{equation}\label{eq:support-crisp}
supp_{crisp}(I) = \frac{count(I)}{N},
\end{equation}
where $count(I)$ is the original support used in~\cite{Agrawal1994} and $N$ is the number of training examples. We have called it $supp_{crisp}$ to differentiate the crisp support used in this stage from the fuzzy support used in the subsequent fuzzy rule generation step (Eq. (\ref{eq:support-fuzzy})). Similarly to Apriori, an itemset is considered as \emph{frequent} if its support equals or exceeds the \emph{support threshold} set by the user. However, in our proposal this threshold depends on the cardinality of the itemset and the number of classes instead of being a fixed number:
\begin{equation}\label{eq:min-support-crisp}
minSupp_{crisp} = \frac{0.025}{|I| \cdot M},
\end{equation}
where $|I|$ is the number of items contained in the itemset, $M$ is the number of classes in the problem, and the value $0.025$ has been set based on empirical results. Unlike Apriori, in this work an itemset might be considered as frequent even if any of its subsets of items are non-frequent. In this manner, only those itemsets having a support lower than the support threshold are discarded. This modification comes from the fact that the minimum support of each itemset depends on the length of the itemset, so that small non-frequent itemsets are penalized more than larger ones. The reason for this adaptation is that the difference between the crisp and the fuzzy supports of larger itemsets is greater than that of smaller ones, which might cause valuable itemsets to be removed if this difference is not minimized.
\item Selection of the most confident itemsets: among the frequent itemsets, another filtering process is carried out based on the \emph{confidence} of the itemsets. In this work, the confidence of an itemset is defined as:
\begin{equation}\label{eq:confidence-crisp}
conf_{crisp}(I) = \frac{\displaystyle \underset{m=1,\ldots,M}{max} \left(countClass(I,y_m)\right)}{count(I)},
\end{equation}
where $countClass(I,y_m)$ counts the number of examples belonging to the class $y_m$ in which the itemset $I$ is present. Similarly to $supp_{crisp}$, we have called it $conf_{crisp}$ to differentiate this confidence from that used in the fuzzy rules generation (Eq. (\ref{eq:confidence-fuzzy})). In order to select the most confident itemsets, the following criteria are used: 
\begin{itemize}
\item For each class, if more than 50\% of the itemsets associated with the class satisfy a certain \emph{confidence threshold} (in this work called $minConf_{crisp}$), those not satisfying the threshold are discarded. Otherwise, itemsets are sorted in descending order by confidence and the bottom 50\% are discarded. This criterion guarantees that at most half of the itemsets are discarded.
\item If there exists a subset of the itemset that is more confident than the itemset itself and fulfills the previous criterion, the itemset is discarded. This means that large itemsets are kept if and only if they provide more discrimination capability than smaller ones.
\end{itemize}
\end{enumerate}
We must remark that any occurrence of a certain itemset is always weighted by the cost associated with the class of the example in which the itemset is present (Eq. (\ref{eq:class-cost})). That is, both the support and the confidence count the occurrences by summing up the cost of the class of each example (same for $N$ in Eq. (\ref{eq:support-crisp})). This procedure is aimed at reducing the impact of the frequency of each class during the learning process.

\subsubsection{Construction of fuzzy rules}\label{sssec:rules-generation}

Based on the most promising frequent itemsets extracted in the previous stage, a fuzzy rule base is created as follows.
\begin{enumerate}[label=(\alph*),leftmargin=2.5\parindent]
\item Conversion from itemsets to candidate rules: every single itemset is transformed into one or more candidate rules. To this end, for a given itemset, the algorithm keeps track of the examples in which the itemset is present and it obtains their class labels. Then, a new candidate rule is generated for each of these classes. Example \ref{ex:itemset-rule} illustrates this conversion.
\begin{example}\label{ex:itemset-rule}
\normalfont
Given the following itemsets that have passed the previous filtering phase:\vspace{5pt}\\
$\{High_2\}, \{Low_1, High_2\}, \{Low_1, High_2, Medium_3\}$.\vspace{5pt}\\
And given the examples that have generated those itemsets:\vspace{5pt}\\
$Low_1, High_2, Medium_3 \rightarrow C_1$\\
$Low_1, High_2, Medium_3 \rightarrow C_2$.\vspace{5pt}\\
We can extract the classes the itemsets belong to ($C_1$ and $C_2$) and generate the corresponding candidate rules:\vspace{5pt}\\
\footnotesize
$\mbox{IF }A_2\mbox{ is }High\mbox{ THEN }C_1$\\
$\mbox{IF }A_2\mbox{ is }High\mbox{ THEN }C_2$\\
$\mbox{IF }A_1\mbox{ is }Low\mbox{ and }A_2\mbox{ is }High\mbox{ THEN }C_1$\\
$\mbox{IF }A_1\mbox{ is }Low\mbox{ and }A_2\mbox{ is }High\mbox{ THEN }C_2$\\
$\mbox{IF }A_1\mbox{ is }Low\mbox{ and }A_2\mbox{ is }High\mbox{ and }A_3\mbox{ is }Medium\mbox{ THEN }C_1$\\
$\mbox{IF }A_1\mbox{ is }Low\mbox{ and }A_2\mbox{ is }High\mbox{ and }A_3\mbox{ is }Medium\mbox{ THEN }C_2$.\\
\end{example}
As we can observe in Example \ref{ex:itemset-rule}, two examples belonging to different classes might produce the same itemsets, and thus the algorithm would generate rules that share the antecedent part and have different consequent. This situation is known as a \emph{conflict} and is resolved in the next step.
\item Computation of rule weights and conflict resolution: for each rule, the matching degree between the rule and all the examples in the training set is computed in order to obtain the rule weight, as shown in Eq. (\ref{eq:rule-weight}). This computation is performed in a distributed fashion by broadcasting all candidate rules across the worker nodes. In this manner, each worker computes only the partial sum of the matching degrees corresponding to the assigned partition. We must point out that this process represents a small fraction of the total computing time of the learning algorithm, since the number of rules considered at this point is generally low. When rule weights are computed, conflicts are resolved by keeping the rule with the largest weight.
\item Filtering and pruning: rules are first filtered based on their support and confidence, which are computed according to Eq. (\ref{eq:support-fuzzy}) and (\ref{eq:confidence-fuzzy}), respectively, reusing the previously computed matching degrees:
\begin{equation}\label{eq:support-fuzzy}
supp_{fuzzy}(R) = \frac{matchClass + matchNotClass}{N}
\end{equation}
\begin{equation}\label{eq:confidence-fuzzy}
conf_{fuzzy}(R) = \frac{matchClass}{matchClass + matchNotClass},
\end{equation}
considering the $matchClass$, $matchNotClass$, and $N$ defined in Eq. (\ref{eq:rule-weight}) and (\ref{eq:support-crisp}), respectively. As described in the extraction of itemsets, $N$ is weighted by the cost of each class. The filtering process consists in removing those rules having a support or a confidence lower than a certain threshold. In this work, the confidence threshold ($minConf_{fuzzy}$) has been set to 0.6. The support threshold is defined as:
\begin{equation}\label{eq:min-support-fuzzy}
minSupp_{fuzzy}(R) = \frac{0.05}{len(R) \cdot M},
\end{equation}
where $len(R)$ and $M$ are the rule length and the number of classes in the problem, respectively, and the value $0.05$ has been set based on empirical results. The usage of the rule length in this computation minimizes the penalizing effect that the product operation has in the matching degree involved in Eq. (\ref{eq:support-fuzzy}) when the number of antecedents increases \cite{Elkano2015,Elkano2016}. 

Finally, a pruning process is carried out to keep only the most confident rules of each class. To this end, all the rules are grouped by class and length and sorted by confidence. Next, for each class $C_m$ with $m \in \{1, 2, ..., M\}$ and rule length $len \in \{1, 2, ..., maxLen\}$, where $maxLen$ is the maximum rule length set by the user, the first $NR(C_m, len)$ rules are taken:
\begin{equation}\label{eq:num-top-rules}
NR(C_m, len) = L \cdot F \cdot prop_{len} \cdot \gamma,
\end{equation}
where $L$ is the number of fuzzy sets per variable and $F$ is the number of variables of the problem. In this work, we set $maxLen = 3$ according to previous studies showing good trade-offs between discrimination capability and complexity \cite{AlcalaFdez2011}. However, for the sake of completeness, we include additional experimental results varying this parameter in the supplementary material. As for $prop_{len}$, it represents the proportion of rules with length $len$ that the rule base should contain after filtering the most confident rules. In this work, we set $prop$ = (0.2, 0.3, 0.5) to prioritize specific rules over general ones. Regarding $\gamma$, it is a hyperparameter that allows the user to set the priority between classification performance and model complexity (in this work $\gamma \in \{2, 4, 8\}$). High values of $\gamma$ cause the algorithm to build more rules and might enhance its classification performance. After filtering the most confident rules, a pruning process is carried out, where those rules containing all the antecedents of a more confident and shorter rule are removed from the rule base. Example \ref{ex:rules-pruning} illustrates this pruning process.
\begin{example}\label{ex:rules-pruning}
\normalfont
Given the following rules:
\begin{equation*}\footnotesize
\begin{split}
&R_1:\mbox{IF }A_1\mbox{ is }Low\mbox{ and }A_2\mbox{ is }High\mbox{ THEN }C_1\\
&R_2:\mbox{IF }A_1\mbox{ is }Low\mbox{ and }A_2\mbox{ is }High\mbox{ and }A_3\mbox{ is }Medium\mbox{ THEN }C_1,\\
\end{split}
\end{equation*}
with $conf_{fuzzy}(R_1)$ = 0.83 and $conf_{fuzzy}(R_2)$ = 0.76, $R_2$ is discarded because $R_1$ is shorter, more confident, and all its antecedents are present in the antecedent part of $R_2$.
\end{example}
\end{enumerate}

Fig. \ref{fig:code-rule-induction} and \ref{fig:diagram-rule-induction} show the pseudo-code of the rule induction algorithm and the four Spark stages launched during the process, respectively. We must highlight two aspects in Fig. \ref{fig:code-rule-induction}:
\begin{itemize}
\item In line 17, candidate rules are grouped by the antecedent part, and thus conflicting rules fall into the same key-value pair. In this manner, \emph{map} and \emph{filter} transformations can compute the support, the confidence, and the weight of all conflicting rules at once.
\item Functions \emph{is\_frequent ()} and \emph{is\_confident ()} check whether the support and the confidence of a given itemset/rule are greater than the corresponding thresholds, respectively.
\end{itemize}
As mentioned in Section \ref{ssec:hadoop-spark}, the shuffling operation occurs only between stages, and thus transformations such as \emph{map} and \emph{filter} run in parallel without communication overhead.
\begin{figure}[!htbp]\scriptsize
	\renewcommand{\algorithmicrequire}{\textbf{Input:}}
	\renewcommand{\algorithmicensure}{\textbf{Output:}}
    \begin{algorithmic}[1]
    \Statex \hspace{-14pt} \textbf{Function} generate\_rule\_base ($TR$)
 	\Require A pre-processed training set $TR$ containing $N$ labeled examples $x_i$.
 	\Ensure A rule base $RB$.
 	\vspace{3pt}
 	\Statex \hspace{-14pt} \textbf{Begin}
 	\vspace{3pt}
 	\State \emph{\textbf{\# 1. Search for the most promising itemsets}}
 	\vspace{3pt}
 	\State \emph{\textbf{\#\hspace{7pt} 1.1 Discretization of the examples}}
 	\vspace{3pt}
 	\State \hspace{10pt} $TR_d \leftarrow TR.map\mbox{ }(x_i \leftarrow discretize\mbox{ }(x_i))$
 	\State \hspace{10pt} $Itemsets \leftarrow TR_d.map\mbox{ }(x^d_i \leftarrow extract\_itemsets\mbox{ }(x^d_i))$
 	\vspace{3pt}
 	\State \emph{\textbf{\#\hspace{7pt} 1.2 Search for frequent itemsets}}
 	\vspace{3pt}
 	\State \hspace{10pt} $SuppConf \leftarrow Itemsets$
 	\Statex \hspace{10pt} $.reduceByKey\mbox{ }(itemset \leftarrow support\_and\_confidence\mbox{ }(itemset))$
 	\State \hspace{10pt} $Itemsets_{Freq} \leftarrow SuppConf.filter\mbox{ }(is\_frequent\mbox{ }(itemset))$
 	\vspace{3pt}
 	\State \emph{\textbf{\#\hspace{7pt} 1.3 Selection of the most confident itemsets}}
 	\vspace{3pt}
 	\State \hspace{10pt} $Itemsets_{Conf} \leftarrow Itemsets_{Freq}.filter\mbox{ }(is\_confident\mbox{ }(itemset))$
 	\State \hspace{10pt} $Itemsets_{Prom} \leftarrow distributed\_pruning\mbox{ }(Itemsets_{Conf})$
 	\vspace{3pt}
 	\State \emph{\textbf{\# 2. Construction of fuzzy rules}}
 	\vspace{3pt}
 	\State \emph{\textbf{\#\hspace{7pt} 2.1 Conversion from itemsets to candidate rules}}
 	\vspace{3pt}
 	\State \hspace{10pt} $Rules_{cand} \leftarrow Itemsets_{Prom}.map\mbox{ }(itemset \leftarrow rule\mbox{ }(itemset))$
 	\State \hspace{10pt} $Rules_{broad} \leftarrow broadcast\mbox{ }(Rules_{cand}.collect\mbox{ }())$
 	\vspace{3pt}
 	\State \emph{\textbf{\#\hspace{7pt} 2.2 Computation of rule weights and conflict resolution}}
 	\vspace{3pt}
 	\State \hspace{10pt} $matchings \leftarrow TR_d.map\mbox{ }(x_i \leftarrow matching\mbox{ }(Rules_{broad}))$
 	\State \hspace{10pt} $SuppConfWght \leftarrow matchings$
 	\Statex \hspace{10pt} $.reduceByKey\mbox{ }(rule \leftarrow support\_confidence\_weight\mbox{ }(rule))$
 	\State \hspace{10pt} $Rules_{no\_conflicts} \leftarrow SuppConfWght$
 	\Statex \hspace{10pt} $.map\mbox{ }(rule \leftarrow resolve\_conflicts\mbox{ }(rule))$
 	\vspace{3pt}
 	\State \emph{\textbf{\#\hspace{7pt} 2.3 Filtering and pruning}}
 	\vspace{3pt}
 	\State \hspace{10pt} $Rules_{freq} \leftarrow Rules_{no\_conflicts}$
 	\Statex \hspace{10pt} $.filter\mbox{ }(rule \leftarrow is\_frequent\mbox{ }(rule))$
 	\State \hspace{10pt} $Rules_{conf} \leftarrow Rules_{freq}.filter\mbox{ }(rule \leftarrow is\_confident\mbox{ }(rule))$
 	\State \hspace{10pt} $Rules \leftarrow distributed\_pruning\mbox{ }(Rules_{conf})$
 	\vspace{3pt}
 	\State \textbf{RETURN} \emph{build\_rule\_base} ($Rules$)
 	\vspace{3pt}
	\Statex \hspace{-14pt} \textbf{End}
	\end{algorithmic}
	\caption{Pseudo-code of the rule induction algorithm.}
	\label{fig:code-rule-induction}
\end{figure}
\begin{figure}[!htbp]
\centering
\includegraphics[scale=0.35]{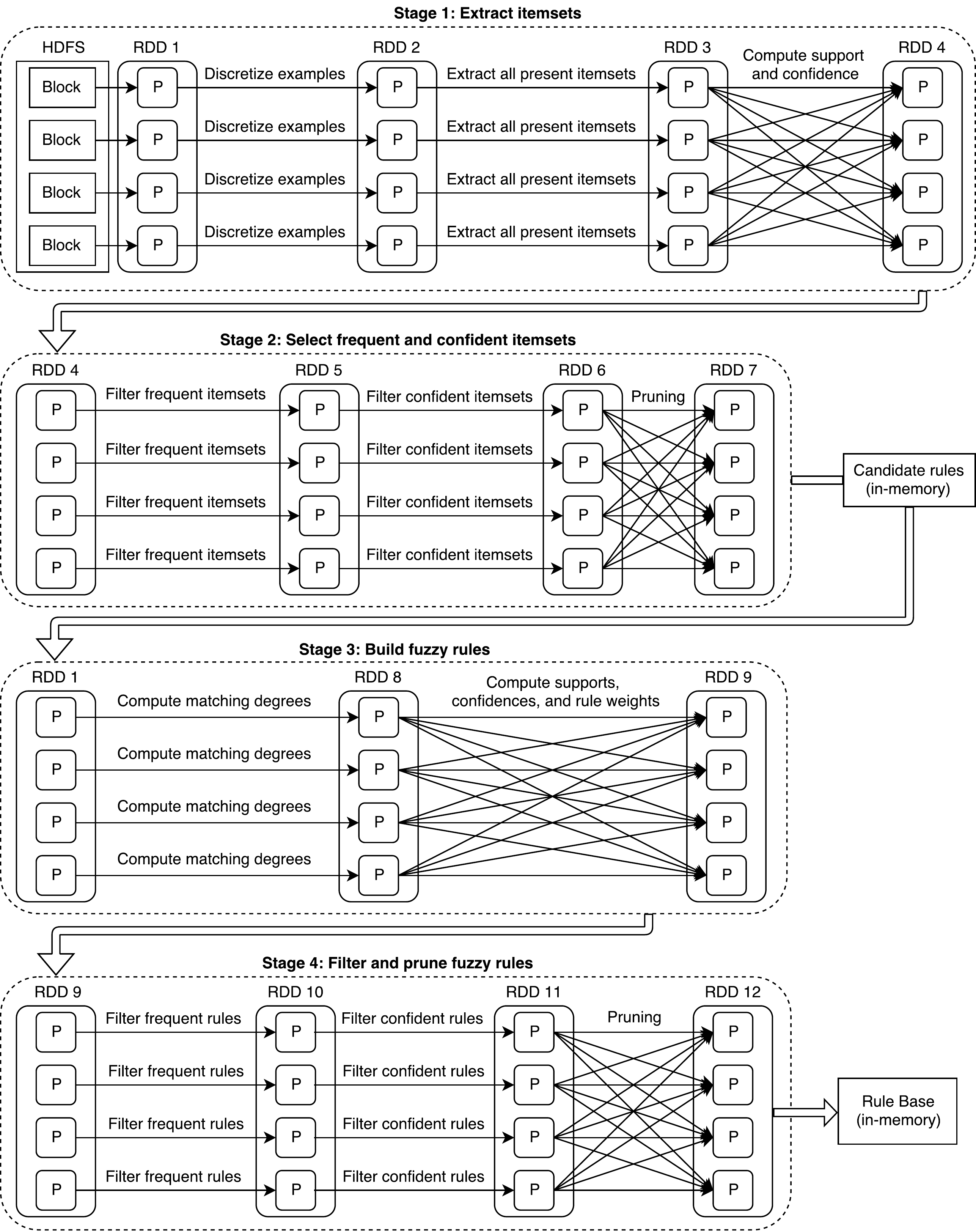}
\caption{Spark stages launched during the rule induction process (P = Partition).}
\label{fig:diagram-rule-induction}
\end{figure}

\subsection{Evolutionary rule selection}\label{ssec:evolutionary}

When the rule base has been created, a rule selection process is carried out in order to obtain a compact and accurate model. To this end, we apply the CHC evolutionary algorithm (EA)~\cite{Eshelman1991} because of its ability to deal with complex search spaces~\cite{Whitley1996} and the good results achieved by this EA in state-of-the-art FRBCSs, such as FARC-HD~\cite{AlcalaFdez2011} or IVTURS~\cite{Sanz2013a}. Unlike other methods that make use of CHC~\cite{Fernandez2016b,Fernandez2017}, we have implemented a new distributed version that performs a global optimization process by evaluating the quality (fitness) of each individual considering the whole training set.

Next, the main features of the CHC algorithm are described:
\begin{itemize}
\item \emph{Coding Scheme}. Each chromosome $C = (c_1, c_2, ..., c_{NR})$ is coded as a binary vector of $NR$ elements, $NR$ being the number of rules contained in the rule base. Each element is associated with a certain rule and determines whether the rule is selected or not. In this manner, $If\mbox{ }c_i = 1\mbox{ }Then\mbox{ }(R_i \in RB)\mbox{ }Else\mbox{ }(R_i \notin RB)$, where $RB$ is the final optimized rule base.
\item \emph{Initial Gene Pool}. To include the initial rule base as a candidate solution, the initial pool is obtained with the first individual setting all genes to 1 and the remaining individuals being generated at random.
\item \emph{Chromosome Evaluation}. Since the goal of our method is to build a compact and accurate model, both the accuracy and the complexity of the rule base need to be considered. To this end, the quality or the \emph{fitness} of a chromosome is determined by the same equation used in FARC-HD:
\begin{equation}\label{eq:fitness}
Fitness(C) = acc - \delta \cdot \frac{NR_{initial}}{NR_{initial}-NR+1},
\end{equation}
where $NR_{initial}$ and $NR$ are the number of rules in the initial and the current rule bases, respectively, and $acc$ is the accuracy obtained by the current model. In order to preserve the descrimination capability for all classes, the accuracy is measured in terms of the geometric mean (GM)~\cite{Barandela2003} defined later in Section \ref{ssec:performance-metrics} (Eq (\ref{eq:geometric-mean})), which is a commonly used metric for imbalanced datasets. 

Since the evaluation of the whole population is the most computationally expensive task in EAs, we have implemented our own distributed version of CHC to parallelize this computation across worker nodes. When computing the fitness of a whole population, all the individuals are sent to every single worker. Then, each worker computes a \emph{partial confusion matrix} for each individual considering only the examples contained in its partitions. Finally, these partial matrices are summed up and the exact accuracy is obtained. In this manner, the fitness of an individual is computed using the whole training set without introducing any approximation. Additionally, in order to avoid repeated computations, we store an RDD containing the pre-computed association degrees of each rule in the initial rule base for all the examples. This allows CHC to use this RDD as a look-up-table when classifying a certain partition. We must point out that the look-up-table is distributed across worker nodes, and thus only the partition assigned to the worker is loaded in the main memory.
\item \emph{Crossover Operator}. The half uniform crossover scheme (HUX) is applied~\cite{Eshelman1993}. The HUX crossover randomly interchanges half of the genes that are different in the parents, ensuring the maximum distance between the offspring and their parents (exploration).
\item \emph{Restarting Approach}. The usage of a restarting approach helps the EA avoid local optima. When the population is restarted, the best chromosome is kept and the remaining are generated at random by keeping a certain percentage ($\Gamma$) of the genes contained in the best chromosome (set by the user).
\end{itemize}

CHC uses an incest prevention mechanism when applying the crossover operator: two parents are crossed only if their hamming distance divided by 2 is greater than a given threshold $D$. This threshold is initialized as the maximum possible distance between two individuals (number of genes in a chromosome) divided by four. When no individuals are added to the next generation, the value of $D$ is decreased by $\varphi$ times its initial value, $\varphi$ being set by the user (in this work 0.01). When $D$ is below zero, and if the maximum number of restarts without improvement has not been reached ($maxRestarts$), the population is restarted.

Although CHC has proven effective in dealing with interpretability-accuracy tradeoffs, it would be interesting to adapt the evolutionary rule selection process to multi-objective frameworks \cite{Ishibuchi2007}. However, this approach is out of the scope of this paper and will be studied in future work.

\section{Experimental framework}\label{sec:experimental-framework}

In this section we present the framework used to develop the experiments carried out in Section \ref{sec:experimental-study}. Firstly, we describe the datasets selected for the experimental study (Section \ref{ssec:datasets}). Next, we introduce the performance and scalability measures used to evaluate the methods (Section \ref{ssec:performance-metrics}). Finally, we show the parameters considered for each method (Section \ref{ssec:parameters}).

\subsection{Datasets} \label{ssec:datasets}

In order to develop the experimental study, we considered 6 Big Data classification problems available at UCI~\cite{Lichman2013} and \fnurl{OpenML}{https://www.openml.org/search?type=data} repositories. Table \ref{tab:datasets} shows the description of the datasets indicating the number of instances (\#Instances), real (R)/integer(I)/categorical(C)/total(T) attributes (\#Attributes), and classes (\#Classes). The names of BNG Australian (BNG), Covertype (COV), HEPMASS (HEPM), and KDDCup1999 (KDD) have been shortened. All the experiments were carried out using a \textit{5-fold stratified cross-validation scheme}. To this end, we randomly split the dataset into five partitions of data, each one containing 20\% of the examples, and we employed a combination of four of them (80\%) to train the system and the remaining one to test it. Therefore, the result of each dataset was computed as the average of the five partitions.

\begin{table}[!htbp]
	\centering
	\caption{Description of the datasets.}
	\label{tab:datasets}
	\renewcommand{\arraystretch}{1.2}
	\begin{tabular}{@{}llllllc@{}}
		\toprule
		Dataset & \#Instances & \multicolumn{4}{@{}c@{}}{\#Attributes} & \#Classes\\
		& & R & I & C & T &\\
		\midrule
		BNG & 1,000,000 & 8    & 6     & 0   & 14 & 2 \\
    	COV & 581,012 & 10    & 0     & 44    & 54 & 7 \\
    	HEPM & 10,500,000 & 28    & 0     & 0     & 28 & 2 \\
    	HIGGS & 11,000,000 & 28    & 0     & 0     & 28 & 2 \\
    	KDD & 4,898,431 & 26    & 0     & 15    & 41 & 5 \\
    	SUSY  & 5,000,000 & 18    & 0     & 0     & 18 & 2 \\
		\bottomrule
	\end{tabular}
\end{table}

\subsection{Performance metrics and scalability measures} \label{ssec:performance-metrics}

The classification performance of the different methods was measured using the accuracy rate ($Acc$), the average accuracy rate per class ($Acc_{Class}$), and the geometric mean ($GM$)~\cite{Barandela2003}, defined as follows:
\begin{equation}\label{eq:accuracy-rate}
Acc = \frac{\displaystyle \sum_{m=1}^{M} TPR_m \cdot N_m}{N},
\end{equation}
\begin{equation}\label{eq:avg-accuracy-rate-class}
Acc_{Class} = \frac{\displaystyle \sum_{m=1}^{M} TPR_m}{M},
\end{equation}
\begin{equation}\label{eq:geometric-mean}
GM = \displaystyle\sqrt[\leftroot{-3}\uproot{3}M]{\displaystyle \prod_{m=1}^{M} TPR_m},
\end{equation}
where $TPR_m$ is the true positive rate of class $C_m$ (proportion of correctly classified examples belonging to class $C_m$), $N_m$ is the number of examples from class $C_m$, and $N$ is the total number of examples. Since some of the datasets considered in the experiments are imbalanced, $Acc_{Class}$ and $GM$ provide more information about the actual discrimination capability than $Acc$.

%In order to give statistical support to the empirical analysis, we carried out some non-parametric tests, as recommended in the specialized literature~\cite{Demsar2006}. More specifically, we used the Wilcoxon signed-ranks test~\cite{Wilcoxon1945} to perform pairwise comparisons, the Aligned Friedman test~\cite{Hodges1962} to check whether there is statistical differences among a group of methods, and the Holm post-hoc test~\cite{Holm1979} to find the algorithms that reject the null hypothesis of equivalence against the selected control method.

Additionally, we measured the scalability of our approach by applying three well-known metrics used to evaluate distributed systems, i.e., speedup, sizeup, and scaleup~\cite{DeWitt1992,Jogalekar2000}.
\begin{itemize}
\item \textit{Speedup}: the data size is kept constant and the number of cores is increased. An ideal distributed algorithm should feature linear speedup, that is, a system with $m$ cores must provide a speedup of $m$. However, in practice a linear speedup is difficult to obtain due to communication and synchronization overhead.
\begin{equation}
Speedup(m)=\frac{\mbox{runtime on 1 core}}{\mbox{runtime on } m \mbox{ cores}}
\end{equation}
\item \textit{Sizeup}: the number of cores is kept constant and the data size is increased. Sizeup measures how much longer it will take to process an $m$-times larger dataset. A linear increase in execution time represents the ideal case.
\begin{equation}
Sizeup(data,m)=\frac{\mbox{runtime for processing } m \cdot data}{\mbox{runtime for processing } data}
\end{equation}
\item \textit{Scaleup}: the ability of a system to run an $m$-times greater job with $m$-times larger system is measured, whose ideal value should be 1 (runtime of the baseline system).
\begin{equation}
\begin{split}
&Scaleup(data,m)=\\ &\frac{\mbox{runtime for processing } data \mbox{ on 1 core}}{\mbox{runtime for processing } m \cdot data \mbox{ on \textit{m} cores}}
\end{split}
\end{equation}

\end{itemize}

\subsection{Methods and parameters setup} \label{ssec:parameters}

We included all the open-source fuzzy classifiers available for Big Data so far, i.e., CHI-BD~\cite{Elkano2017}, Chi-Spark-RS~\cite{Fernandez2017}, and FMDT/FBDT\cite{Segatori2018b}. Chi-FRBCS-BigData~\cite{Lopez2015} was not included in the comparisons because CHI-BD showed better performance in terms of accuracy and complexity in~\cite{Elkano2017}. Although MLlib provides a non-fuzzy version of decision trees for Big Data, we did not include this method because Segatori et al. already carried out a comparative study in which FBDT/FMDT outperformed MLlib decision trees \cite{Segatori2018b}. Table \ref{tab:parameters} shows the parameters considered for each method throughout the experiments. In all cases, we set the values suggested by the authors in the original papers.

\begin{table}[!htbp]\footnotesize
	\centering
	\caption{Parameters used for each method.}
	\label{tab:parameters}
	\begin{tabular}{@{}ll@{}}
		\toprule
		\textit{Algorithm} & \textit{Parameters} \\
		\midrule
		& \#Fuzzy sets per variable = 5\\
		& Inference = winning rule \\
		CFM-BD & Rule weight = PCF-CS\\
		(rule induction) & $maxLen$ = 3; $prop$ = (0.2, 0.3, 0.5) \\
		& $minConf_{crisp}$ = 0.7; $minConf_{fuzzy}$ = 0.6 \\
		& $\gamma \in \{2, 4, 8\}$ \\
		\cmidrule(r){2-2}
		& \#Individuals = 50; \#Evaluations = 10,000 \\
		CFM-BD & $maxRestarts$ = 3; $D$ = $NR_{initial}$ / 4 \\
		(rule selection) & $\delta$ = 0.15; $\Gamma$ = 0.35; $\varphi$ = 0.01 \\
		\midrule
		& Impurity = entropy; T-norm = product \\
		FMDT & $maxBins$ = 32; $maxDepth$ ($\beta$) = 5 \\
		& $\gamma$ = 0.1\%; $\phi$ = 0.02 $\cdot$ $N$; $\lambda$ = 10$^{-4}\cdot N$ \\
		\midrule
		& Impurity = entropy; T-norm = product \\
		FBDT & $maxBins$ = 32; $maxDepth$ ($\beta$) $\in \{5, 10, 15\}$\\
		& $\gamma$ = 0.1\%; $\phi$ = 1; $\lambda$ = 1 \\
		\midrule
		& \#Fuzzy sets per variable = 3\\
		& Inference = winning rule \\
		CHI-BD & Rule weight = PCF-CS\\
		(cost-sensitive) & Number of rule subsets = 4\\
		& Minimum \#occurrences \\ 
		& for frequent subsets = 10\\
		& Maximum \#rules per reducer = 400,000\\
		\midrule
		& \#Fuzzy sets per variable = 3\\
		Chi-Spark-RS & Inference = winning rule \\
		(cost-sensitive) & Rule weight = PCF-CS; T-norm = product\\
		& \#Individuals = 50; \#Evaluations = 1,000 \\
		& $\alpha$ (for fitness) = 0.7\\
		\bottomrule
	\end{tabular}
\end{table}

In the case of CFM-BD, there is an extra boolean parameter called $cost\_sensitive$ that enables/disables the cost-sensitive mode. When it is off, the cost associated with each class ($cost(y_m)$) is set to 1, so that the learning algorithm ignores the frequency of each class. Also, different parts of the algorithm are adapted accordingly:
\begin{itemize}
\item The pruning of itemsets described in Section \ref{ssec:rule-induction} considers all the itemsets at once regardless of their class. 
\item Similarly, the most confident fuzzy rules are extracted without considering their classes, so that rules are grouped only by length. Consequently, $NR(C_m, len)$ (Eq. (\ref{eq:num-top-rules})) is replaced with $NR(len)$ (Eq. (\ref{eq:num-top-rules-no-cost-sensitive})) by adding the number of classes $M$:
\begin{equation}\label{eq:num-top-rules-no-cost-sensitive}
NR(len) = L \cdot F \cdot prop_{len} \cdot \gamma \cdot M.
\end{equation}
\item In addition to the rule induction process, the evolutionary optimization is also modified by replacing the $GM$ with the $Acc$ to focus on improving accuracy rate.
\end{itemize}
\noindent This way, the user can turn on/off the cost-sensitive mode of CFM-BD by simply setting the parameter $cost\_sensitive$. When optimizing the $Acc$, the non cost-sensitive version performs better in general, while the best $Acc_{Class}$ and $GM$ are obtained when using the cost-sensitive version. The $Acc$ reported in this work corresponds to the non cost-sensitive CFM-BD, while the $Acc_{Class}$ and $GM$ correspond to the cost-sensitive version.

Additionally, we include two versions of CFM-BD: CFM-BD and CFM-BD$^{L}$. The former corresponds to the original method introduced in Section \ref{sec:cfm-bd}, while the latter is a lightweight mode that omits the evolutionary rule selection process. Therefore, the only difference between these two versions is that CFM-BD$^{L}$ gets rid of the third stage described in Section \ref{ssec:evolutionary}. The original CFM-BD provides more accurate and compact models than CFM-BD$^{L}$, since the latter is meant to achieve a good trade-off among classification performance, model complexity, and execution time.

Regarding the cluster used for running the algorithms, it is composed of 6 slave nodes and a master node connected via 1Gb/s Ethernet LAN network. Half of the slave nodes have 2 Intel Xeon E5-2620 v3 processors at 2.4 GHz (3.2 GHz with Turbo Boost) with 12 virtual cores in each one (where 6 of them are physical). The other half are equipped with 2 Intel Xeon E5-2620 v2 processors at 2.1 GHz with the same number of cores as the previous ones. The master node is composed of an Intel Xeon E5-2609 processor with 4 physical cores at 2.4 GHz. All slave nodes are equipped with 64 GB of RAM memory, while the master works with 32 GB of RAM memory. With respect to the storage specifications, all nodes use Hard Disk Drives featuring a read/write performance of 128 MB/s. The entire cluster runs on top of CentOS 6.5 + Apache Hadoop 2.6.0 + Apache Spark 2.0.2.

Except for FMDT/FBDT, the number of partitions/cores used for the execution of the algorithms equals the maximum number supported by our cluster, i.e., 128. In the case of FMDT/FBDT, we found that using more than 24 cores had a negative impact on runtimes when setting the configuration recommended by the authors. Thus, the number of cores used for FMDT/FBDT was 24. In all cases, we assigned 4 cores to every single executor in order to ensure full HDFS write throughput while minimizing memory replication overhead (e.g. broadcast variables).

\section{Experimental study}\label{sec:experimental-study}

In this section we describe the empirical study carried out to assess the performance of the proposed method (CFM-BD), which consists of two parts:
\begin{enumerate}
\item We tested CFM-BD in six Big Data classification problems and compared its performance with that provided by four state-of-the-art fuzzy classifiers (FMDT/FBDT~\cite{Segatori2018b}, Chi-Spark-RS~\cite{Fernandez2017}, and CHI-BD~\cite{Elkano2017}) (Section \ref{ssec:performance}). More specifically, the performance of these methods was evaluated in terms of classification accuracy, model complexity, and runtimes. %Additionally, non-parametric statistical tests were applied in order to provide statistical support to the analysis.
\item We assessed the scalability of our approach with three well-known metrics used to evaluate distributed systems, i.e., \emph{speedup}, \emph{sizeup}, and \emph{scaleup}~\cite{DeWitt1992,Jogalekar2000} (Section \ref{ssec:scalability}).
\end{enumerate}

\subsection{Classification performance and complexity}\label{ssec:performance}

Tables \ref{tab:classification-performance} and \ref{tab:rules} show the classification performance of each method and the model complexities in terms of the number of rules (\#rules), average rule length ($\overline{RL}$), average number of fuzzy sets per variable ($\overline{FS}$), and total rule length ($\overline{TRL}=\mbox{\#rules}\cdot\overline{RL}\cdot\overline{FS}$), respectively. The second column indicates the $\gamma$ used for CFM-BD (Eq. (\ref{eq:num-top-rules})) and the maximum depth considered for FBDT and FMDT ($\beta$). In order to replicate the configurations suggested in~\cite{Segatori2018b}, the maximum depth used for FMDT is always 5. As we can observe, larger $\gamma$'s do not imply better classification performance in the case of CFM-BD, while deeper trees generally provide more accurate models in FBDT. Notice that the best classification performance achieved for each dataset is highlighted in boldface (Table \ref{tab:classification-performance}).

\begin{table}[!htbp]\scriptsize
\centering
\renewcommand\tabcolsep{3pt}
\renewcommand{\arraystretch}{1}
\caption{Classification performance of each method.}
\label{tab:classification-performance}
\begin{tabular}{@{}llcccccc@{}}
	\toprule
	\multicolumn{8}{c}{\textbf{Accuracy rate} \% ($Acc$)}\\
	\\
    Dataset & $\gamma$; $\beta$ & CFM-BD & CFM-BD$^{L}$ & FBDT & FMDT & Chi-Spark-RS & CHI-BD\\
	\midrule
	& 2; 5  & 86.45 & 85.42 & 78.83 & 80.23 &       &  \\
    BNG & 4; 10 & 86.05 & 85.31 & 80.17 &       & 75.19 & 84.21 \\
          & 8; 15 & \textbf{86.54} & 85.32 & 80.17 &       &       &  \\
          \midrule
          & 2; 5  & 72.67 & 69.67 & 69.31 & \textbf{94.28} &       &  \\
    COV & 4; 10 & 72.61 & 69.75 & 76.30 &       & -     & 51.69 \\
          & 8; 15 & 72.49 & 70.18 & 82.87 &       &       &  \\
          \midrule
          & 2; 5  & 90.60 & 89.15 & 90.61 & -     &       &  \\
    HEPM & 4; 10 & 90.75 & 89.23 & 91.15 &       & -     & - \\
          & 8; 15 & 90.75 & 89.20 & \textbf{91.40} &       &       &  \\
          \midrule
          & 2; 5  & 65.11 & 62.19 & 66.39 & 71.54 &       &  \\
    HIGGS & 4; 10 & 68.19 & 63.23 & 70.60 &       & -     & 58.54 \\
          & 8; 15 & 68.47 & 63.48 & \textbf{72.23} &       &       &  \\
          \midrule
          & 2; 5  & 99.07 & 98.79 & 99.88 & \textbf{99.99} &       &  \\
    KDD & 4; 10 & 98.81 & 98.79 & \textbf{99.99} &       & -     & 99.65 \\
          & 8; 15 & 98.82 & 98.80 & \textbf{99.99} &       &       &  \\
          \midrule
          & 2; 5  & 77.53 & 75.88 & 77.31 & 79.29 &       &  \\
    SUSY & 4; 10 & 78.41 & 76.13 & 79.09 &       & -     & 64.89 \\
          & 8; 15 & 78.85 & 76.45 & \textbf{79.70} &       &       &  \\
	\toprule
	\multicolumn{8}{c}{\textbf{Average accuracy rate \% per class} ($Acc_{Class}$)}\\ 
	\\
	Dataset & $\gamma$; $\beta$ & CFM-BD & CFM-BD$^{L}$ & FBDT & FMDT & Chi-Spark-RS & CHI-BD\\
	\midrule
	& 2; 5  & 86.20 & 85.02 & 77.44 & 78.96 &       &  \\
    BNG & 4; 10 & 86.32 & 84.99 & 78.81 &       & 75.35 & 85.02 \\
          & 8; 15 & \textbf{86.36} & 85.00 & 78.84 &       &       &  \\
          \midrule
          & 2; 5  & 62.09 & 59.25 & 44.59 & \textbf{87.25} &       &  \\
    COV & 4; 10 & 64.74 & 60.88 & 62.17 &       & -     & 66.73 \\
          & 8; 15 & 67.98 & 63.08 & 75.65 &       &       &  \\
          \midrule
          & 2; 5  & 90.62 & 89.15 & 90.61 & -     &       &  \\
    HEPM & 4; 10 & 90.69 & 89.23 & 91.15 &       & -     & - \\
          & 8; 15 & 90.74 & 89.07 & \textbf{91.40} &       &       &  \\
          \midrule
          & 2; 5  & 67.90 & 64.63 & 66.23 & 71.43 &       &  \\
    HIGGS & 4; 10 & 68.90 & 65.29 & 70.47 &       & -     & 58.48 \\
          & 8; 15 & 68.87 & 65.30 & \textbf{72.09} &       &       &  \\
          \midrule
          & 2; 5  & 95.59 & 93.39 & 59.47 & 92.15 &       &  \\
    KDD & 4; 10 & 95.68 & 93.00 & 89.84 &       & -     & 83.94 \\
          & 8; 15 & \textbf{95.90} & 94.29 & 90.60 &       &       &  \\
          \midrule
          & 2; 5  & 76.21 & 74.53 & 76.57 & 78.59 &       &  \\
    SUSY & 4; 10 & 77.13 & 74.88 & 78.35 &       & -     & 62.42 \\
          & 8; 15 & 78.27 & 75.38 & \textbf{79.01} &       &       &  \\
	\toprule
	\multicolumn{8}{c}{\textbf{Geometric mean} ($GM$)}\\
	\\
	Dataset & $\gamma$; $\beta$ & CFM-BD & CFM-BD$^{L}$ & FBDT & FMDT & Chi-Spark-RS & CHI-BD\\
	\midrule
	& 2; 5  & .8620 & .8484 & .7685 & .7847 &       &  \\
    BNG & 4; 10 & .8632 & .8480 & .7825 &       & .7534 & .8483 \\
          & 8; 15 & \textbf{.8636} & .8481 & .7832 &       &       &  \\
          \midrule
          & 2; 5  & .5917 & .5448 & .2861 & \textbf{.8680} &       &  \\
    COV & 4; 10 & .6240 & .5667 & .5738 &       & -     & .6419 \\
          & 8; 15 & .6602 & .5930 & .7417 &       &       &  \\
          \midrule
          & 2; 5  & .9061 & .8908 & .9059 & -     &       &  \\
    HEPM & 4; 10 & .9069 & .8917 & .9114 &       & -     & - \\
          & 8; 15 & .9073 & .8904 & \textbf{.9139} &       &       &  \\
          \midrule
          & 2; 5  & .6787 & .6447 & .6619 & .7140 &       &  \\
    HIGGS & 4; 10 & .6887 & .6528 & .7043 &       & -     & .5847 \\
          & 8; 15 & .6882 & .6530 & \textbf{.7205} &       &       &  \\
          \midrule
          & 2; 5  & .9529 & .9307 & .0000 & .9076 &       &  \\
    KDD & 4; 10 & .9556 & .9265 & .8833 &       & -     & .7594 \\
          & 8; 15 & \textbf{.9578} & .9415 & .8880 &       &       &  \\
          \midrule
          & 2; 5  & .7617 & .7452 & .7603 & .7815 &       &  \\
    SUSY & 4; 10 & .7712 & .7487 & .7787 &       & -     & .5524 \\
          & 8; 15 & .7816 & .7537 & \textbf{.7858} &       &       &  \\
    \bottomrule
\end{tabular}
\end{table}
\begin{table*}[!htbp]\tiny
\centering
\renewcommand\tabcolsep{3pt}
\renewcommand{\arraystretch}{1.1}
\caption{Complexity of each method.}
\label{tab:rules}
\begin{tabular}{@{}llrrrrrrrrrrrrrrrrrrrrrrrrrr@{}}
	\toprule
    Dataset & $\gamma$; $\beta$ & \multicolumn{4}{c}{CFM-BD} & \multicolumn{4}{c}{CFM-BD$^{L}$} & \multicolumn{4}{c}{FBDT} & \multicolumn{4}{c}{FMDT} & \multicolumn{4}{c}{Chi-Spark-RS} & \multicolumn{4}{c}{CHI-BD}\\ \\
    & & \multicolumn{1}{c}{\#rules} & \multicolumn{1}{c}{$\overline{RL}$} & \multicolumn{1}{c}{$\overline{FS}$} & \multicolumn{1}{c}{$\overline{TRL}$} & \multicolumn{1}{c}{\#rules} & \multicolumn{1}{c}{$\overline{RL}$} & \multicolumn{1}{c}{$\overline{FS}$} & \multicolumn{1}{c}{$\overline{TRL}$} & \multicolumn{1}{c}{\#rules} & \multicolumn{1}{c}{$\overline{RL}$} & \multicolumn{1}{c}{$\overline{FS}$} & \multicolumn{1}{c}{$\overline{TRL}$} & \multicolumn{1}{c}{\#rules} & \multicolumn{1}{c}{$\overline{RL}$} & \multicolumn{1}{c}{$\overline{FS}$} & \multicolumn{1}{c}{$\overline{TRL}$} & \multicolumn{1}{c}{\#rules} & \multicolumn{1}{c}{$\overline{RL}$} & \multicolumn{1}{c}{$\overline{FS}$} & \multicolumn{1}{c}{$\overline{TRL}$} & \#rules & $\overline{RL}$ & $\overline{FS}$ & \multicolumn{1}{c}{$\overline{TRL}$} \\
	\cmidrule{1-1}\cmidrule(r){2-2}\cmidrule(r){3-6}\cmidrule(r){7-10}\cmidrule(r){11-14}\cmidrule(r){15-18}\cmidrule(r){19-22}\cmidrule(r){23-26}
		& 2; 5  & 5     & 1.97  & 5.00  & 51    & 241   & 2.51  & 5.00  & 3,018 & 32    & 5.00  & 6.04  & 967   & 83,044 & 3.02  & 6.04  & 1,515,513 &       &       &       &       &       &       &       &  \\
    BNG & 4; 10 & 8     & 1.89  & 5.00  & 72    & 455   & 2.58  & 5.00  & 5,858 & 666   & 9.69  & 6.04  & 38,992 &       &       &       &       & 6,493 & 14.00 & 3.00  & 272,706 & 8,720 & 14.00 & 3.00  & 366,240 \\
          & 8; 15 & 9     & 1.81  & 5.00  & 81    & 857   & 2.63  & 5.00  & 11,266 & 6,302 & 14.22 & 6.04  & 541,615 &       &       &       &       &       &       &       &       &       &       &       &  \\
          \cmidrule{1-1}\cmidrule(r){2-2}\cmidrule(r){3-6}\cmidrule(r){7-10}\cmidrule(r){11-14}\cmidrule(r){15-18}\cmidrule(r){19-22}\cmidrule(r){23-26}
          & 2; 5  & 164   & 2.70  & 5.00  & 2,219 & 2,158 & 2.78  & 5.00  & 30,032 & 31    & 4.97  & 18.84 & 2,901 & 268,677 & 3.39  & 18.84 & 17,183,934 &       &       &       &       &       &       &       &  \\
    COV & 4; 10 & 565   & 2.86  & 5.00  & 8,076 & 3,908 & 2.87  & 5.00  & 56,090 & 779   & 9.87  & 18.84 & 144,824 &       &       &       &       & -     & -     & -     & -     & 5,135 & 54.00 & 3.00  & 831,870 \\
          & 8; 15 & 1,576 & 2.94  & 5.00  & 23,133 & 6,763 & 2.92  & 5.00  & 98,876 & 8,723 & 14.36 & 18.84 & 2,359,402 &       &       &       &       &       &       &       &       &       &       &       &  \\
          \cmidrule{1-1}\cmidrule(r){2-2}\cmidrule(r){3-6}\cmidrule(r){7-10}\cmidrule(r){11-14}\cmidrule(r){15-18}\cmidrule(r){19-22}\cmidrule(r){23-26}
          & 2; 5  & 7     & 1.00  & 5.00  & 34    & 462   & 2.49  & 5.00  & 5,762 & 30    & 4.93  & 22.20 & 3,286 & -     & -     & -     & -     &       &       &       &       &       &       &       &  \\
    HEPM & 4; 10 & 9     & 1.35  & 5.00  & 61    & 867   & 2.55  & 5.00  & 11,066 & 681   & 9.84  & 22.20 & 148,858 &       &       &       &       & -     & -     & -     & -     & -     & -     & -     & - \\
          & 8; 15 & 13    & 1.71  & 5.00  & 115   & 1,693 & 2.60  & 5.00  & 22,006 & 13,805 & 14.77 & 22.20 & 4,525,926 &       &       &       &       &       &       &       &       &       &       &       &  \\
          \cmidrule{1-1}\cmidrule(r){2-2}\cmidrule(r){3-6}\cmidrule(r){7-10}\cmidrule(r){11-14}\cmidrule(r){15-18}\cmidrule(r){19-22}\cmidrule(r){23-26}
          & 2; 5  & 22    & 2.41  & 5.00  & 268   & 440   & 2.61  & 5.00  & 5,747 & 32    & 5.00  & 13.01 & 2,081 & 6,414,575 & 3.25  & 13.01 & 271,561,578 &       &       &       &       &       &       &       &  \\
    HIGGS & 4; 10 & 32    & 2.47  & 5.00  & 398   & 873   & 2.62  & 5.00  & 11,438 & 849   & 9.89  & 13.01 & 109,142 &       &       &       &       & -     & -     & -     & -     & 666,068 & 28.00 & 3.00  & 55,949,712 \\
          & 8; 15 & 51    & 2.60  & 5.00  & 664   & 1,486 & 2.71  & 5.00  & 20,168 & 17,390 & 14.79 & 13.01 & 3,345,832 &       &       &       &       &       &       &       &       &       &       &       &  \\
          \cmidrule{1-1}\cmidrule(r){2-2}\cmidrule(r){3-6}\cmidrule(r){7-10}\cmidrule(r){11-14}\cmidrule(r){15-18}\cmidrule(r){19-22}\cmidrule(r){23-26}
          & 2; 5  & 44    & 2.28  & 5.00  & 507   & 1,639 & 2.58  & 5.00  & 21,153 & 24    & 4.83  & 15.93 & 1,848 & 8,042 & 3.18  & 15.93 & 407,509 &       &       &       &       &       &       &       &  \\
    KDD & 4; 10 & 227   & 2.60  & 5.00  & 2,948 & 3,110 & 2.64  & 5.00  & 41,033 & 171   & 8.80  & 15.93 & 24,010 &       &       &       &       & -     & -     & -     & -     & 5,646 & 41.00 & 3.00  & 694,458 \\
          & 8; 15 & 740   & 2.74  & 5.00  & 10,138 & 5,343 & 2.76  & 5.00  & 73,610 & 369   & 11.89 & 15.93 & 69,806 &       &       &       &       &       &       &       &       &       &       &       &  \\
          \cmidrule{1-1}\cmidrule(r){2-2}\cmidrule(r){3-6}\cmidrule(r){7-10}\cmidrule(r){11-14}\cmidrule(r){15-18}\cmidrule(r){19-22}\cmidrule(r){23-26}
          & 2; 5  & 11    & 1.02  & 5.00  & 58    & 325   & 2.43  & 5.00  & 3,955 & 32    & 5.00  & 22.60 & 3,616 & 5,225,134 & 3.68  & 22.60 & 435,136,493 &       &       &       &       &       &       &       &  \\
    SUSY & 4; 10 & 19    & 1.60  & 5.00  & 151   & 606   & 2.53  & 5.00  & 7,668 & 718   & 9.78  & 22.60 & 158,646 &       &       &       &       & -     & -     & -     & -     & 9,505 & 18.00 & 3.00  & 513,270 \\
          & 8; 15 & 21    & 1.76  & 5.00  & 187   & 1,169 & 2.58  & 5.00  & 15,080 & 11,054 & 14.63 & 22.60 & 3,654,337 &       &       &       &       &       &       &       &       &       &       &       &  \\
	\bottomrule
\end{tabular}
\end{table*}

Before analyzing and comparing each method, we must mention that the implementation of Chi-Spark-RS available at \fnurl{GitHub}{https://github.com/aFdezHilario/Chi-Spark-RS} does not support multi-class problems, and thus we could not run this algorithm on KDD and COV. Besides, this method was not able to tackle HIGGS, HEPMASS, and SUSY within a period of 48 hours, and hence no results are given on these datasets for this method. Consequently, we will not consider Chi-Spark-RS in the analysis of the results. Also, FMDT and CHI-BD ran out of memory on HEPMASS and we were not able to extract any result on this dataset.

Next, we analyze and compare the performance of CFM-BD in terms of classification performance and model complexity for each dataset:
\begin{itemize}
\item BNG: in this case CFM-BD clearly surpasses the rest of methods, improving the accuracy of FBDT, FMDT, and Chi-Spark-RS by more than 6\% ($Acc$) and 8\% ($Acc_{Class}$ and $GM$), while building less than 10 rules. In the case of CHI-BD, the difference in classification performance is not that large (though CFM-BD is still more accurate). However, CHI-BD builds nearly 9K rules composed of 14 antecedents, yielding a $\overline{TRL}$ 3-7K times greater than CFM-BD's.
\item COV: FMDT stands out from the rest of methods but generates a model with a $\overline{TRL}$ value of 17M (0,7-8K times the $\overline{TRL}$ of CFM-BD). As for FBDT, the only model that provides a comparable $\overline{TRL}$ with respect to CFM-BD (FBDT$_{\beta=5}$) is not able to maintain classification performance. In fact, when it comes to $Acc_{Class}$ and $GM$, FBDT requires a minimum depth of 10 to provide a more accurate model than CFM-BD's, yielding a $\overline{TRL}$ 19 and 6 times greater than that of CFM-BD$_{\gamma=4}$ and CFM-BD$_{\gamma=8}$, respectively. The improvement obtained by deeper trees in FBDT suggests that COV requires more than 3 antecedents per rule to achieve state-of-the-art classification performance, which might explain the performance loss in CFM-BD. Regarding CHI-BD, it is unable to provide competitive $Acc$ but achieves comparable $Acc_{Class}$ and $GM$ with respect to CFM-BD. However, CFM-BD provides a $\overline{TRL}$ 40-400 times smaller than CHI-BD's.
\item HEPM: only deep FBDTs are able to obtain slightly better classification performance than CFM-BD (by less than \%1), but using a $\overline{TRL}$ 1-100K times greater than CFM-BD's. The rest of models ran out of memory during the execution.
\item HIGGS: FBDT and FMDT performs better than CFM-BD only with considerably more complex models (200-20K times the $\overline{TRL}$ of CFM-BD). With respect to CHI-BD, CFM-BD clearly outperforms this method by up to 10\%.
\item KDD: all methods perform well in terms of $Acc$ (98-99\%), but CFM-BD stands out from all the rest when it comes to $Acc_{Class}$ and $GM$. As for model complexities, FBDT is the only method that is able to extract fewer rules than CFM-BD. However, the rules generated by FBDT are 2-5 times longer and use 3 times more fuzzy sets, leading to larger $\overline{TRL}$s.
\item SUSY: CFM-BD achieves competitive classification performance with respect to the best method (FBDT$_{\beta=15}$), using 1K and 10 times fewer rules and antecedents, respectively. Furthermore, the accuracy of CFM-BD with 21 rules ($\overline{TRL}=187$) is quite similar to that of FMDT with 5M rules ($\overline{TRL}=400M$).
\end{itemize}
We must remark that some datasets have certain properties that affect the classification performance and/or model complexity of CFM-BD:
\begin{itemize}
\item If the great majority of features are categorical (COV has 44 out of 54), CFM-BD is not able to take full advantage of fuzzy logic. This type of feature also causes the algorithm to build more rules when there exist many possible values.
\item Unlike FBDT and FMDT, CFM-BD is designed to discriminate all classes, and thus multi-class problems such as COV and KDD require more rules to face the greater overlap that generally exists among the examples of different classes. As a result, shallow FBDTs might yield fewer rules than CFM-BD, though the $\overline{TRL}$ of such models is still larger than CFM-BD's.
\end{itemize}

The execution times of each method are shown in Table \ref{tab:runtimes}. As expected, the original evolutionary version of CFM-BD is slower than the rest. Although we prioritized the interpretability of the model over the execution time, we have proposed a non-evolutionary lightweight version of CFM-BD$^{L}$ that is much faster than CFM-BD, as shown in Table \ref{tab:runtimes}. Of course, this version sacrifices both accuracy and interpretability to reduce execution times.

\begin{table}[!htbp]\scriptsize
\centering
\renewcommand\tabcolsep{3.5pt}
\renewcommand{\arraystretch}{1.1}
\caption{Runtime (s) of each method.}
\label{tab:runtimes}
\begin{tabular}{@{}llrrrrrr@{}}
	\toprule
    Dataset & $\gamma$; $\beta$ & CFM-BD & CFM-BD$^{L}$ & FBDT & FMDT & Chi-Spark-RS & CHI-BD\\
	\midrule
	& 2; 5  & 3,079 & 33    & 65    & 84    &       &  \\
    BNG   & 4; 10 & 5,170 & 32    & 84    &       & 1,740 & 77 \\
          & 8; 15 & 12,692 & 33    & 108   &       &       &  \\
          \midrule
          & 2; 5  & 9,592 & 496   & 29    & 113   &       &  \\
    COV & 4; 10 & 13,406 & 481   & 49    &       & -     &  \\
          & 8; 15 & 26,386 & 428   & 87    &       &       &  \\
          \midrule
          & 2; 5  & 11,805 & 2,646 & 325   & -     &       &  \\
    HEPM  & 4; 10 & 28,313 & 2,654 & 391   &       & -     & - \\
          & 8; 15 & 60,167 & 2,399 & 623   &       &       &  \\
          \midrule
          & 2; 5  & 22,961 & 3,252 & 309   & 5,238 &       &  \\
    HIGGS & 4; 10 & 37,167 & 3,300 & 474   &       & -     & 9,658 \\
          & 8; 15 & 57,762 & 3,015 & 716   &       &       &  \\
          \midrule
          & 2; 5  & 29,064 & 582   & 105   & 77    &       &  \\
    KDD   & 4; 10 & 56,341 & 599   & 157   &       & -     & 81 \\
          & 8; 15 & 116,584 & 499   & 186   &       &       &  \\
          \midrule
          & 2; 5  & 6,483 & 397   & 148   & 1,392 &       &  \\
    SUSY  & 4; 10 & 12,547 & 403   & 235   &       & -     & 103 \\
          & 8; 15 & 21,822 & 351   & 364   &       &       &  \\
	\bottomrule
\end{tabular}
\end{table}

Overall, CFM-BD achieves state-of-the-art discrimination capabilities with respect to the best performing algorithms (FBDT and FMDT), while providing much simpler models that can be interpreted. The models of CFM-BD generally consist of a few rules composed of less than 3 antecedents when keeping $\gamma$ below 4, whereas other methods generate thousands of rules containing many antecedents and linguistic labels. Additionally, the experimental results have revealed that the non-evolutionary rule induction algorithm of CFM-BD (CFM-BD$^{L}$) provides a good trade-off among classification performance, model complexity, and execution time. We must mention that the model and time complexities of CFM-BD shown in this paper correspond to the cost-sensitive mode. Anyway, the complexities are similar regardless of this feature.

\subsection{Scalability}\label{ssec:scalability}

\begin{figure*}[!htbp]
        \centering
        \subfloat[Speedup]{\scalebox{0.7}{\includegraphics{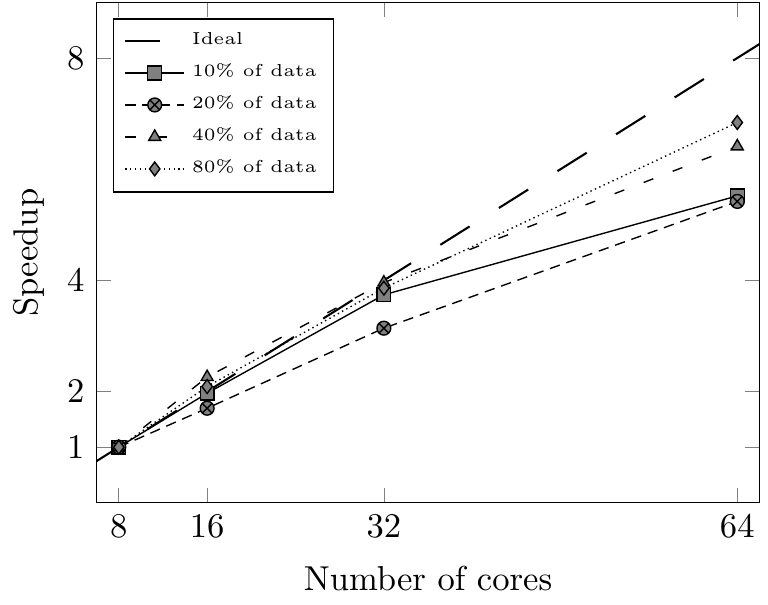}}\label{fig:speedup-candidates}}\hspace{10pt}
        \subfloat[Sizeup]{\scalebox{0.7}{\includegraphics{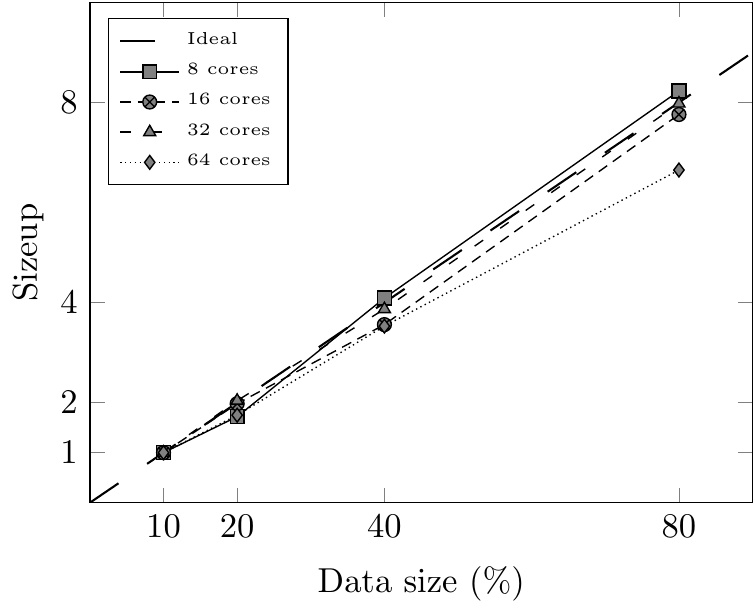}}\label{fig:sizeup-candidates}}\hspace{10pt}
        \subfloat[Scaleup]{\scalebox{0.7}{\includegraphics{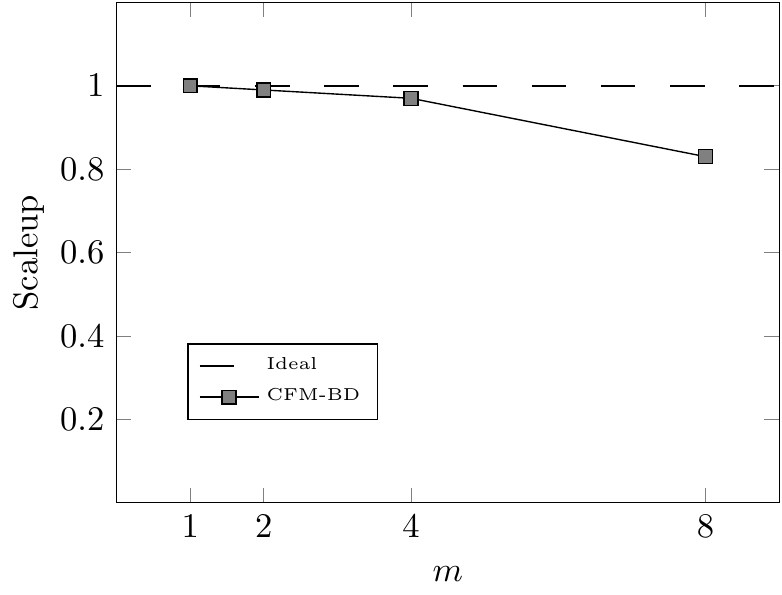}}\label{fig:scaleup-candidates}}
    \caption{Efficiency of the rule induction process on HIGGS.}
    \label{fig:efficiency-candidates}
\end{figure*}

\begin{figure*}[!htbp]
        \centering
        \subfloat[Speedup]{\scalebox{0.7}{\includegraphics{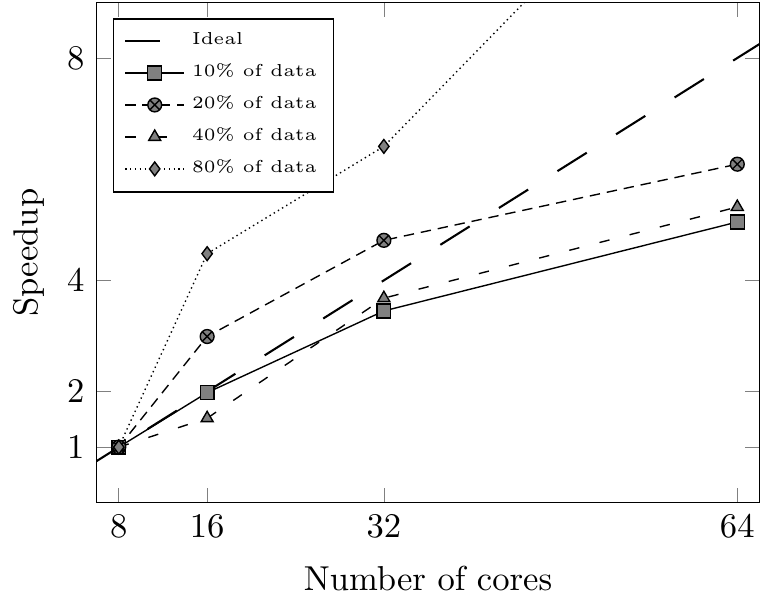}}\label{fig:speedup-total}}\hspace{10pt}
        \subfloat[Sizeup]{\scalebox{0.7}{\includegraphics{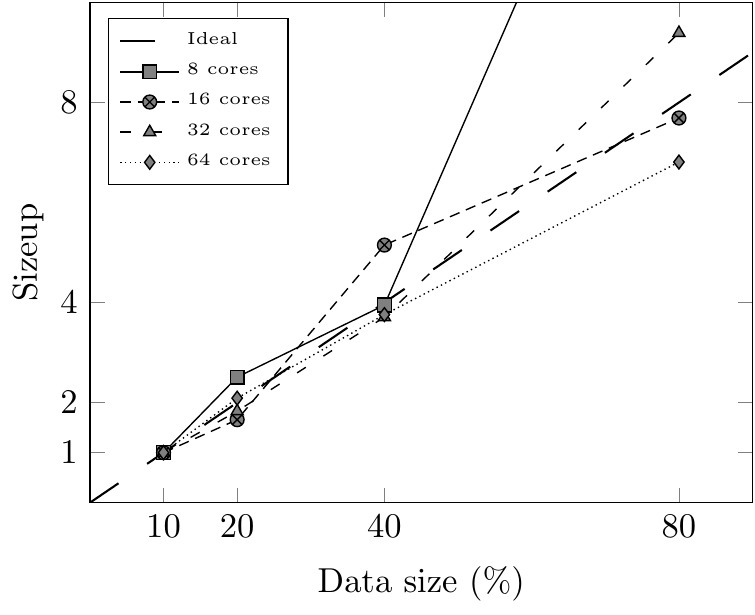}}\label{fig:sizeup-total}}\hspace{10pt}
        \subfloat[Scaleup]{\scalebox{0.7}{\includegraphics{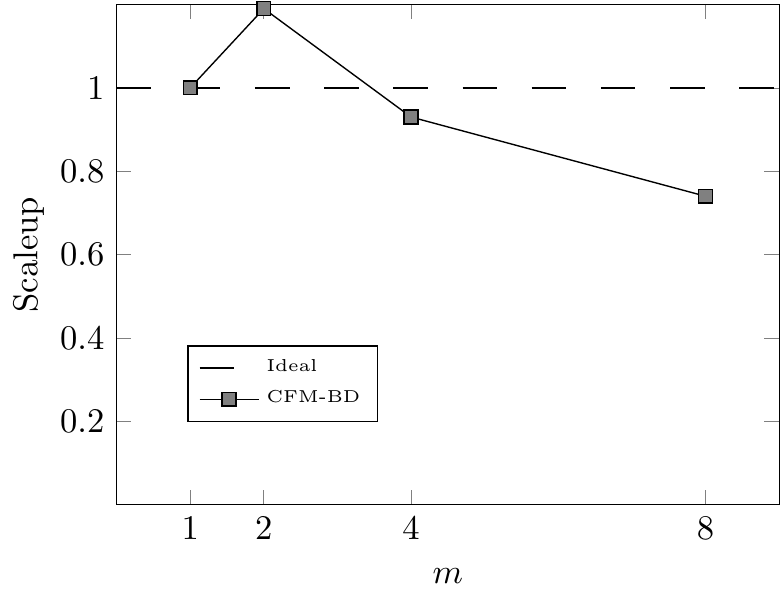}}\label{fig:scaleup-total}}
    \caption{Efficiency of the whole learning algorithm on HIGGS.}
    \label{fig:efficiency-total}
\end{figure*}

Finally, we study the efficiency of our approach in terms of speedup, sizeup, and scaleup~\cite{DeWitt1992,Jogalekar2000} by testing CFM-BD on several reduced versions of HIGGS and varying the number of cores used for the execution. More specifically, we take 8 cores and 10\% of HIGGS as the baseline case ($m$ = 1) and we gradually double both the number of cores and the data size (maintaining the original class distribution), until 64 cores and 80\% of HIGGS. This way, for each number of cores (8, 16, 32, 64) we run the model using 10\%, 20\%, 40\%, and 80\% of data. In addition to the total execution time, we test the efficiency of the most critical stage of the learning algorithm: the rule induction process. With the runtimes obtained in these executions we build the matrices shown in Table \ref{tab:runtimes-scalability}, which are used to compute the speedup, sizeup, and scaleup (Fig. \ref{fig:efficiency-candidates} and \ref{fig:efficiency-total}). All the executions haven been carried out using the configuration that leads to the best trade-off between classification performance and model complexity ($\gamma=4$).
\begin{table}[!htbp]\scriptsize
  \centering
  \caption{Runtime (s) of CFM-BD on HIGGS.}
  \renewcommand{\arraystretch}{1.1}
  \renewcommand\tabcolsep{7pt}
    \begin{tabular}{@{}ccrrrr@{}}
    \toprule
    Stage & Data size & 8 cores & 16 cores & 32 cores & 64 cores \\
    \midrule
    & 10\% & 1,486 & 754 & 397 & 269 \\
    Rule induction & 20\% & 2,549 & 1,495 & 811 & 470 \\
    & 40\% & 6,084 & 2,687 & 1,537 & 949 \\
    & 80\% & 12,226 & 5,850 & 3,169 & 1,788 \\
    \midrule
    & 10\% & 27,445 & 13,872 & 7,949 & 5,434 \\
    Whole learning & 20\% & 68,846 & 23,000 & 14,576 & 11,296 \\
    algorithm & 40\% & 108,492 & 71,415 & 29,505 & 20,410 \\
    & 80\% & 478,612 & 106,728 & 74,642 & 37,018 \\
    \bottomrule
    \end{tabular}%
  \label{tab:runtimes-scalability}%
\end{table}%

As we can observe in Fig. \ref{fig:efficiency-candidates} and \ref{fig:efficiency-total}, both the rule induction process and the whole learning algorithm reveal nearly linear speedup and sizeup and are able to maintain the scalability above 0.83 and 0.74, respectively. However, when it comes to the whole learning process, Fig. \ref{fig:efficiency-total} shows large variations when using 8 cores. The reason behind this behavior is the pre-computation step performed before the evolutionary optimization. Since the matching degrees of the rules do not change during the rule selection process, a distributed look-up table stores the matching degrees between the rules and the examples before launching the evolutionary algorithm. When this table does not fit into a cached RDD, the efficiency of the evolutionary optimization drastically drops.

\section{Concluding remarks}\label{sec:conclusions}

In this paper we have presented a new distributed FRBCS for Big Data classification problems named CFM-BD. The majority of fuzzy classifiers designed for Big Data so far are based on adaptations or extensions of existing learning algorithms. None of these approaches has been specifically designed from scratch to provide a good trade-off between accuracy and interpretability in Big Data problems.

The goal of this work was to build compact and interpretable models that achieve competitive classification performance. To this end, we have proposed a new rule induction process inspired by CHI-BD~\cite{Elkano2017} and Apriori~\cite{Agrawal1994} algorithms called CFM-BD. Although it employs concepts introduced by these two methods, CFM-BD does not adapt, extend, or combine any of them. Instead, it applies a new learning algorithm composed of three stages: preprocessing, rule induction, and global evolutionary rule selection. All these stages have been specifically designed for Big Data from scratch in order to process the whole training set in a distributed fashion and perform global optimization tasks that do not introduce any approximation error. As a result, CFM-BD always provides exactly the same model regardless of the degree of parallelism used for the execution.

The experimental results show that the models generated by CFM-BD are significantly simpler than the rest. In terms of the number of rules, CFM-BD generally builds a few rules composed of less than 3 antecedents, while other methods generate thousands of rules containing many antecedents and linguistic labels. The only exception are the shallowest models of FBDT, which usually build less than 32 rules. However, these models are generally less accurate than CFM-BD and do not achieve state-of-the-art classification performance. Moreover, the rules constructed by FBDT are more complex than those of CFM-BD, since the number of fuzzy sets used for each variable depends on the variable itself and is often greater than 10. In CFM-BD, each variable is modeled with 5 fuzzy sets that are adjusted to the actual distribution of the variable, obtaining accurate and interpretable rules. In addition to interpretability, CFM-BD has been shown to be competitive in terms of classification performance, providing state-of-the-art discrimination capability.

\ifCLASSOPTIONcaptionsoff
  \newpage
\fi

\bibliographystyle{IEEEtran}

\vspace{80pt}
\begin{IEEEbiography}[{\includegraphics[width=1in,height=1.25in,clip,keepaspectratio]{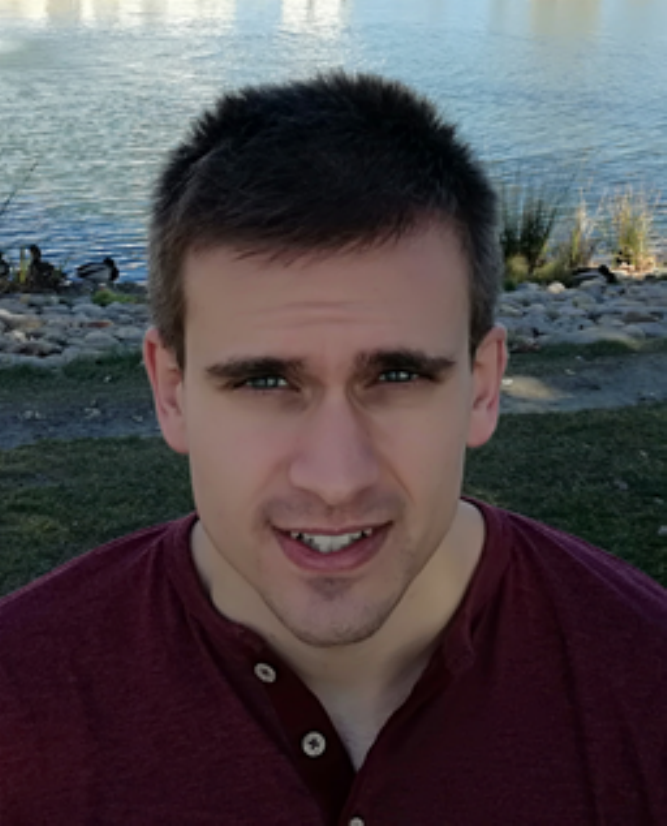}}]{Mikel Elkano}
received his PhD degree in Computer Science from the Public University of Navarre (2018), where he is currently working as a postdoctoral researcher at the Institute of Smart Cities. Previously, he received his BSc in Computer Science from the same university (2014) and continued his studies into a MSc in Psychobiology and Cognitive Neuroscience at the Autonomous University of Barcelona (2016). His research interests include deep learning, neuroscience, fuzzy systems, and big data.
\end{IEEEbiography}
\vspace{-40pt}
\begin{IEEEbiography}[{\includegraphics[width=1in,height=1.25in,clip,keepaspectratio]{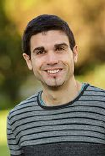}}]{Jos\'e Antonio Sanz}
received his MSc (2008) and PhD (2011) degrees in Computer Ccience from the Public University of Navarra. He is currently an Assistant Professor at the Department of Statistics, Computer Science and Mathematics at the Public University of Navarra. He is the author of 32 published original articles in international journals and more than 45 contributions to conferences. He received the best paper award in the FLINS 2012 international conference and the Pepe Mill\'a award in 2014. His research interests include fuzzy techniques for classification problems, interval-valued fuzzy sets, genetic fuzzy systems and medical applications using soft computing techniques.
\end{IEEEbiography}
\vspace{-40pt}
\begin{IEEEbiography}[{\includegraphics[width=1in,height=1.25in,clip,keepaspectratio]{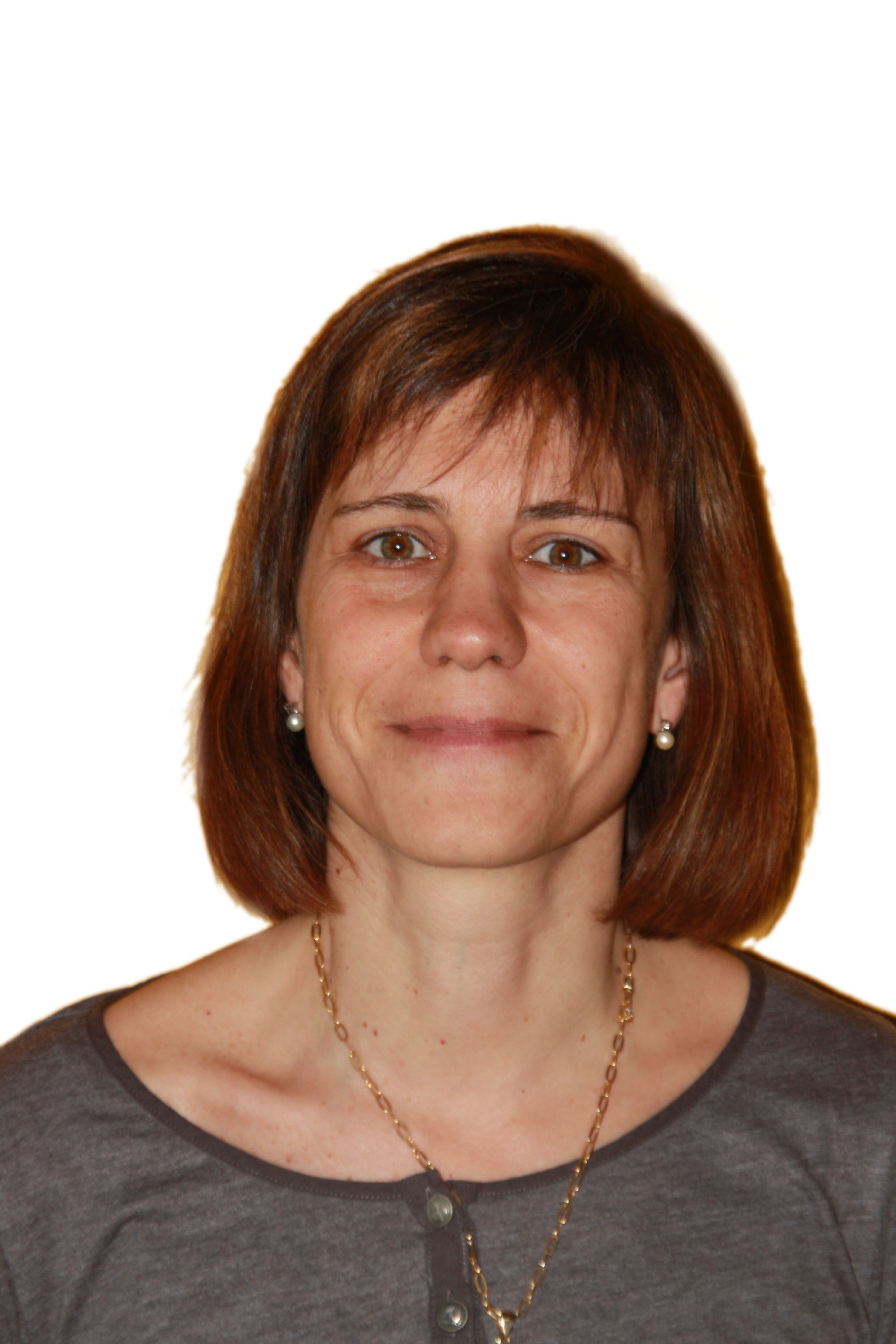}}]{Edurne Barrenechea}
received her MSc (1990) and PhD (2005) degrees in Computer Science from the University of the Basque Country and the Public University of Navarre, respectively. She worked for a private company until 2001 when she joined the Public University of Navarre. Currently she is an Associate Professor at the Department of Statistics, Computer Science and Mathematics. Her publications comprise more than 55 papers in JCR and around 20 book chapters. Her research interests include fuzzy techniques for image processing, fuzzy sets theory, interval type-2 fuzzy sets theory and applications, decision making, and medical and industrial applications of soft computing techniques. She is a Board Member of the Spanish Society for Artificial Intelligence (AEPIA) and belongs to the Institute of Smart Cities of her University.
\end{IEEEbiography}
\vspace{-40pt}
\begin{IEEEbiography}[{\includegraphics[width=1in,height=1.25in,clip,keepaspectratio]{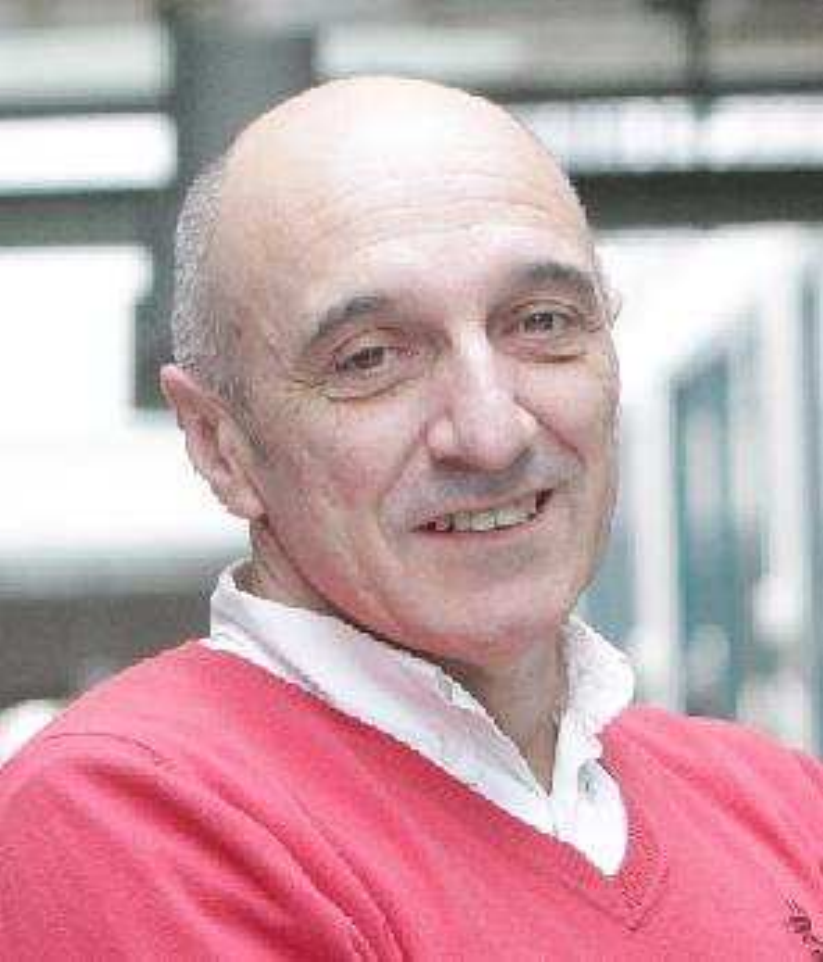}}]{Humberto Bustince}
received his BSc in Physics from the University of Salamanca (1983) and his PhD in Mathematics from the Public University of Navarra (1994). He is a Full Professor of Computer Science and Artificial Intelligence at the Public University of Navarra, where he leads the Artificial Intelligence and Approximate Reasoning Research Group. He has led more than 10 R+D public-funded research projects and authored more than 210 works, according to Web of Science. He is an Associated Editor of the IEEE Transactions on Fuzzy Systems Journal and a member of the editorial board of the Journals Fuzzy Sets and Systems, Information Fusion, International Journal of Computational Intelligence Systems and Journal of Intelligent \& Fuzzy Systems.
\end{IEEEbiography}
\vspace{-40pt}
\begin{IEEEbiography}[{\includegraphics[width=1in,height=1.25in,clip,keepaspectratio]{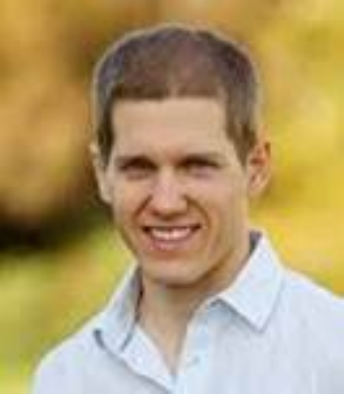}}]{Mikel Galar}
received his MSc (2009) and PhD (2012) degrees in Computer Science from the Public University of Navarra. He is currently an Associate Professor at the Department of Statistics, Computer Science and Mathematics at the Public University of Navarre. He is the author of 35 published original articles in international journals and more than 50 contributions to conferences. He is also reviewer of more than 35 international journals and received the extraordinary prize for his PhD thesis from the Public University of Navarre and the 2013 IEEE Transactions on Fuzzy System Outstanding Paper Award for the paper “A New Approach to Interval-Valued Choquet Integrals and the Problem of Ordering in Interval-Valued Fuzzy Set Applications” (bestowed in 2016).
\end{IEEEbiography}



\section*{Supplementary material}

# Supplementary Material

# CFM-BD: a distributed rule induction algorithm for building Compact Fuzzy Models in Big Data classification problems

Mikel Elkano, Jose Sanz, Edurne Barrenechea,
Humberto Bustince, *Senior Member, IEEE*, Mikel Galar, *Member, IEEE*

## Abstract

This is a supplementary document for "CFM-BD: a distributed rule induction algorithm for building Compact Fuzzy Models in Big Data classification problems". In this document, we provide additional experimental results to explain how the preprocessing stage and the maximum rule length ($maxLen$) affect the performance of the lightweight version of CFM-BD (CFM-BD$^L$). We have considered CFM-BD$^L$ instead of CFM-BD because the stages where the preprocessing algorithm and the hyperparameter $maxLen$ are directly involved correspond to the rule induction process (as explained in the paper, CFM-BD$^L$ is equivalent to running CFM-BD without the evolutionary stage). In this analysis, we measure classification performance, model complexity, and runtimes.

## APPENDIX A
### IMPACT OF THE PRE-PROCESSING STAGE

As explained in the paper, the pre-processing stage of CFM-BD plays an important role in the discrimination capability of linguistic labels (fuzzy sets). When this stage is removed from the algorithm (CFM-BD$^{No\_Pre}$), the fuzzy sets are built by simply distributing a number of triangular membership functions across the attribute domain in a uniform fashion. Consequently, neither the shape nor the position of the fuzzy sets is adjusted to the real distribution of training data, leading to worse classification performance (Table I). In terms of model complexity and runtimes (Tables IIa and IIb), we can observe a decrease in the number of rules when getting rid of the pre-processing stage. This behavior suggests that many fuzzy sets might be delimited by low-density regions of the input space when they are built on the original training set (Fig. 1). As a result, the number of fuzzy sets that can be used for distinguishing the great majority of data points drops dramatically, leading to less itemsets and reduced discrimination capabilities.

## APPENDIX B
### IMPACT OF THE MAXIMUM RULE LENGTH

In order to analyze the impact of the maximum rule length ($maxLen$) we tested three different values: $maxLen = 2$ (CFM-BD$_2^L$) with $prop = (0.2, 0.8)$, $maxLen = 3$ (CFM-BD$_3^L$) with $prop = (0.2, 0.3, 0.5)$, and $maxLen = 4$ (CFM-BD$_4^L$) with $prop = (0.1, 0.2, 0.3, 0.4)$. Some methods were not able to train a model within a period of 24 hours or ran out of memory on certain datasets (marked as '-'). A detailed description of the CPU and memory resources used for the experiments can be found in the paper.

In principle, one could expect better classification performance when increasing the value of $maxLen$ as a result of the larger search space explored to obtain all the existing itemsets containing $maxLen$ items at most. Of course, the algorithm requires more computing resources as the value of $maxLen$ is increased, and thus the most suitable configuration would be that providing a good trade-off. According to the experimental results, gradually increasing $maxLen$ from 2 to 4 seems to be useless in terms of classification performance in most of the cases (Table III) and apparently does not worth the cost of extra computing resources required

TABLE I: Classification performance of each method.

| Dataset | $\gamma$ | Accuracy rate % ($Acc$) | | Accuracy rate % per class ($Acc_{Class}$) | | Geometric mean ($GM$) | |
|---|---|---|---|---|---|---|---|
| | | CFM-BD$^L_{No\_Pre}$ | CFM-BD$^L$ | CFM-BD$^L_{No\_Pre}$ | CFM-BD$^L$ | CFM-BD$^L_{No\_Pre}$ | CFM-BD$^L$ |
| BNG | 2 | 84.96 | **85.42** | 84.88 | **85.02** | .8468 | **.8484** |
| | 4 | 84.96 | 85.31 | 84.88 | 84.99 | .8468 | .8480 |
| | 8 | 84.94 | 85.32 | 84.90 | 85.00 | .8470 | .8481 |
| COV | 2 | 68.14 | 69.67 | 56.17 | 59.25 | .5072 | .5448 |
| | 4 | 68.20 | 69.75 | 55.48 | 60.88 | .4787 | .5667 |
| | 8 | 68.72 | **70.18** | 56.49 | **63.08** | .4973 | **.5930** |
| HEPM | 2 | 88.24 | 89.15 | 88.05 | 89.15 | .8804 | .8908 |
| | 4 | 87.71 | **89.23** | 87.39 | **89.23** | .8739 | **.8917** |
| | 8 | 87.33 | 89.20 | 87.34 | 89.07 | .8733 | .8904 |
| HIGGS | 2 | 54.06 | 62.19 | 52.03 | 64.63 | .4720 | .6447 |
| | 4 | 54.06 | 63.23 | 52.03 | 65.29 | .4720 | .6528 |
| | 8 | 54.06 | **63.48** | 52.03 | **65.30** | .4720 | **.6530** |
| KDD | 2 | 99.12 | 98.79 | 90.62 | 93.39 | .8972 | .9307 |
| | 4 | 99.12 | 98.79 | 91.26 | 93.00 | .9043 | .9265 |
| | 8 | **99.14** | 98.80 | 89.95 | **94.29** | .8901 | **.9415** |
| SUSY | 2 | 66.97 | 75.88 | 67.81 | 74.53 | .6744 | .7452 |
| | 4 | 66.27 | 76.13 | 67.64 | 74.88 | .6716 | .7487 |
| | 8 | 66.29 | **76.45** | 67.65 | **75.38** | .6718 | **.7537** |

TABLE II: Runtimes and model complexity of each method.

(a) Complexity of each method.

| Dataset | $\gamma$ | CFM-BD$^L_{No\_Pre}$ | | | CFM-BD | | |
|---|---|---|---|---|---|---|---|
| | | #rules | $\overline{RL}$ | $\overline{TRL}$ | #rules | $\overline{RL}$ | $\overline{TRL}$ |
| BNG | 2 | 228 | 2.56 | 2,915 | 241 | 2.51 | 3,018 |
| | 4 | 402 | 2.65 | 5,335 | 455 | 2.58 | 5,858 |
| | 8 | 637 | 2.78 | 8,853 | 857 | 2.63 | 11,266 |
| COV | 2 | 2,121 | 2.79 | 29,611 | 2,158 | 2.78 | 30,032 |
| | 4 | 3,816 | 2.87 | 54,762 | 3,908 | 2.87 | 56,090 |
| | 8 | 6,371 | 2.92 | 93,091 | 6,763 | 2.92 | 98,876 |
| HEPM | 2 | 430 | 2.56 | 5,509 | 462 | 2.49 | 5,762 |
| | 4 | 841 | 2.60 | 10,945 | 867 | 2.55 | 11,066 |
| | 8 | 1,627 | 2.62 | 21,289 | 1,693 | 2.60 | 22,006 |
| HIGGS | 2 | 9 | 3.00 | 132 | 440 | 2.61 | 5,747 |
| | 4 | 9 | 3.00 | 132 | 873 | 2.62 | 11,438 |
| | 8 | 9 | 3.00 | 132 | 1,486 | 2.71 | 20,168 |
| KDD | 2 | 1,674 | 2.57 | 21,551 | 1,639 | 2.58 | 21,153 |
| | 4 | 3,213 | 2.62 | 42,045 | 3,110 | 2.64 | 41,033 |
| | 8 | 5,552 | 2.72 | 75,463 | 5,343 | 2.76 | 73,610 |
| SUSY | 2 | 275 | 2.57 | 3,535 | 325 | 2.43 | 3,955 |
| | 4 | 480 | 2.67 | 6,409 | 606 | 2.53 | 7,668 |
| | 8 | 827 | 2.81 | 11,604 | 1,169 | 2.58 | 15,080 |

(b) Runtime (s) of each method.

| Dataset | $\gamma$ | CFM-BD$^L_{No\_Pre}$ | CFM-BD$^L$ |
|---|---|---|---|
| BNG | 2 | 18 | 33 |
| | 4 | 18 | 32 |
| | 8 | 18 | 33 |
| COV | 2 | 99 | 496 |
| | 4 | 98 | 481 |
| | 8 | 101 | 428 |
| HEPM | 2 | 875 | 2,646 |
| | 4 | 868 | 2,654 |
| | 8 | 881 | 2,399 |
| HIGGS | 2 | 895 | 3,252 |
| | 4 | 900 | 3,300 |
| | 8 | 905 | 3,015 |
| KDD | 2 | 221 | 582 |
| | 4 | 222 | 599 |
| | 8 | 217 | 499 |
| SUSY | 2 | 38 | 397 |
| | 4 | 37 | 403 |
| | 8 | 36 | 351 |

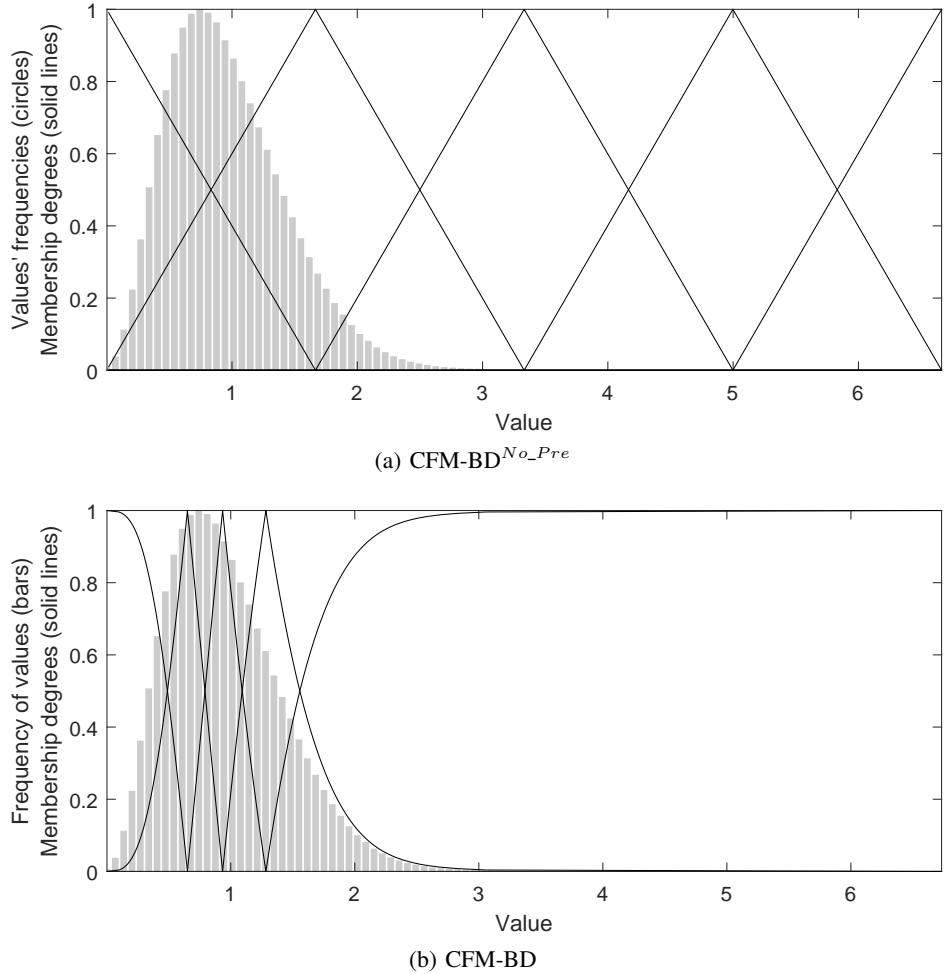

(a) CFM-BD$^{No\_Pre}$

(b) CFM-BD

Fig. 1: Fuzzy sets built by CFM-BD$^{No\_Pre}$ and CFM-BD for the variable $R$ on SUSY.

by increased model complexities and runtimes (Tables IV and V). These findings suggest that the most suitable value of $maxLen$ for CFM-BD$^L$ might be 2 instead of 3 (default value used in the paper). However, since the goal of this work is to design an algorithm for building compact fuzzy models, we have focused on the evolutionary version of CFM-BD and set $maxLen = 3$ to allow the evolutionary algorithm (which is not included in CFM-BD$^L$) to explore more solutions. In order to perform fair comparisons of both methods (CFM-BD and CFM-BD$^L$) in the experiments carried out in the paper, we set the same value of $maxLen$ for both.

TABLE III: Classification performance of each method.

| Dataset | $\gamma$ | Accuracy rate % ($Acc$) | | | Accuracy rate % per class ($Acc_{Class}$) | | | Geometric mean ($GM$) | | |
|---|---|---|---|---|---|---|---|---|---|---|
| | | CFM-BD$_2^L$ | CFM-BD$_3^L$ | CFM-BD$_4^L$ | CFM-BD$_2^L$ | CFM-BD$_3^L$ | CFM-BD$_4^L$ | CFM-BD$_2^L$ | CFM-BD$_3^L$ | CFM-BD$_4^L$ |
| BNG | 2 | 85.31 | **85.42** | 85.33 | 84.98 | **85.02** | 84.99 | .8479 | **.8484** | .8480 |
| | 4 | 85.31 | 85.31 | **85.42** | 84.98 | 84.99 | 84.99 | .8479 | .8480 | .8480 |
| | 8 | 85.31 | 85.32 | 85.31 | 84.98 | 85.00 | 84.99 | .8479 | .8480 | .8481 |
| COV | 2 | 69.63 | 69.67 | - | 57.12 | 59.25 | - | .5136 | .5448 | - |
| | 4 | 69.63 | 69.75 | - | 57.12 | 60.88 | - | .5136 | .5667 | - |
| | 8 | 69.63 | **70.18** | - | 57.12 | **63.08** | - | .5136 | **.5930** | - |
| HEPM | 2 | **89.23** | 89.15 | - | **89.23** | 89.15 | - | **.8917** | .8908 | - |
| | 4 | 89.11 | **89.23** | - | 88.99 | **89.23** | - | .8896 | **.8917** | - |
| | 8 | 88.71 | 89.20 | - | 88.71 | 89.07 | - | .8868 | .8904 | - |
| HIGGS | 2 | 63.14 | 62.19 | - | 65.07 | 64.63 | - | .6507 | .6447 | - |
| | 4 | 63.27 | 63.23 | - | 64.99 | 65.29 | - | .6499 | .6528 | - |
| | 8 | 63.27 | **63.48** | - | 64.99 | **65.30** | - | .6499 | **.6530** | - |
| KDD | 2 | 98.79 | 98.79 | 98.79 | 94.46 | 93.39 | 92.02 | .9435 | .9307 | .9158 |
| | 4 | 98.79 | 98.79 | 98.79 | 94.43 | 93.58 | 93.73 | .9431 | .9265 | .9326 |
| | 8 | 98.79 | **98.80** | 98.79 | 94.43 | 94.29 | 94.63 | .9431 | .9415 | **.9449** |
| SUSY | 2 | 76.20 | 75.88 | 75.89 | 74.96 | 74.53 | 74.44 | .7495 | .7452 | .7444 |
| | 4 | **76.48** | 76.13 | 76.01 | 75.46 | 74.88 | 74.57 | .7545 | .7487 | .7456 |
| | 8 | 76.40 | 76.45 | 76.20 | **75.64** | 75.38 | 75.06 | **.7561** | .7537 | .7505 |

TABLE IV: Complexity of each method.

| Dataset | $\gamma$ | CFM-BD$_2^L$ | | | CFM-BD$_3^L$ | | | CFM-BD$_4^L$ | | |
|---|---|---|---|---|---|---|---|---|---|---|
| | | #rules | $\overline{RL}$ | $\overline{TRL}$ | #rules | $\overline{RL}$ | $\overline{TRL}$ | #rules | $\overline{RL}$ | $\overline{TRL}$ |
| BNG | 2 | 201 | 1.91 | 1,916 | 241 | 2.51 | 3,018 | 269 | 3.08 | 4,138 |
| | 4 | 300 | 1.94 | 2,908 | 455 | 2.58 | 5,858 | 521 | 3.15 | 8,198 |
| | 8 | 300 | 1.94 | 2,910 | 857 | 2.63 | 11,266 | 1,014 | 3.20 | 16,210 |
| COV | 2 | 502 | 1.98 | 4,968 | 2,158 | 2.78 | 30,032 | - | - | - |
| | 4 | 503 | 1.98 | 4,980 | 3,908 | 2.87 | 56,056 | - | - | - |
| | 8 | 515 | 1.98 | 5,096 | 6,763 | 2.92 | 98,876 | - | - | - |
| HEPM | 2 | 400 | 1.89 | 3,786 | 462 | 2.49 | 5,762 | - | - | - |
| | 4 | 737 | 1.94 | 7,158 | 867 | 2.55 | 11,066 | - | - | - |
| | 8 | 1,444 | 1.97 | 14,230 | 1,693 | 2.60 | 22,006 | - | - | - |
| HIGGS | 2 | 374 | 1.98 | 3,708 | 440 | 2.61 | 5,747 | - | - | - |
| | 4 | 418 | 1.99 | 4,148 | 873 | 2.62 | 11,438 | - | - | - |
| | 8 | 418 | 1.99 | 4,148 | 1,486 | 2.71 | 20,168 | - | - | - |
| KDD | 2 | 1,175 | 1.97 | 11,568 | 1,639 | 2.58 | 21,153 | 1,806 | 3.17 | 28,595 |
| | 4 | 1,276 | 1.97 | 12,576 | 3,110 | 2.64 | 41,033 | 3,645 | 3.20 | 58,400 |
| | 8 | 1,275 | 1.97 | 12,562 | 5,343 | 2.76 | 73,610 | 6,934 | 3.28 | 113,790 |
| SUSY | 2 | 279 | 1.86 | 2,596 | 325 | 2.43 | 3,955 | 360 | 2.99 | 5,382 |
| | 4 | 520 | 1.93 | 5,014 | 606 | 2.53 | 7,668 | 680 | 3.11 | 10,565 |
| | 8 | 969 | 1.96 | 9,504 | 1,169 | 2.58 | 15,080 | 1,320 | 3.17 | 20,940 |

TABLE V: Runtime (s) of each method.

| Dataset | $\gamma$ | CFM-BD$_2^L$ | CFM-BD$_3^L$ | CFM-BD$_4^L$ |
|---|---|---|---|---|
| BNG | 2 | 9 | 33 | 68 |
| | 4 | 9 | 32 | 67 |
| | 8 | 8 | 33 | 68 |
| COV | 2 | 23 | 496 | - |
| | 4 | 22 | 481 | - |
| | 8 | 22 | 428 | - |
| HEPM | 2 | 93 | 2,646 | - |
| | 4 | 89 | 2,654 | - |
| | 8 | 91 | 2,399 | - |
| HIGGS | 2 | 96 | 3,252 | - |
| | 4 | 95 | 3,300 | - |
| | 8 | 97 | 3,015 | - |
| KDD | 2 | 19 | 582 | 3,097 |
| | 4 | 19 | 599 | 2,953 |
| | 8 | 18 | 499 | 3,097 |
| SUSY | 2 | 27 | 397 | 1,166 |
| | 4 | 27 | 403 | 1,231 |
| | 8 | 27 | 351 | 1,151 |



\end{document}